\documentclass{ieeeaccess}
\usepackage{cite}
\usepackage{amsmath,amssymb,amsfonts}
\usepackage{pifont}
\usepackage{multirow}
\usepackage{algorithmic}
\usepackage{graphicx}
\usepackage[caption=false,font=footnotesize,labelfont=sf,textfont=sf]{subfig}
\usepackage{textcomp}
\def\BibTeX{{\rm B\kern-.05em{\sc i\kern-.025em b}\kern-.08em
    T\kern-.1667em\lower.7ex\hbox{E}\kern-.125emX}}

\newcommand{\cmark}{\ding{51}}%
\newcommand{\xmark}{\ding{53}}%

\begin{document}
\history{Date of publication xxxx 00, 0000, date of current version xxxx 00, 0000.}
\doi{10.1109/ACCESS.2023.0322000}

\title{PAtt-Lite: Lightweight Patch and Attention MobileNet for Challenging Facial Expression Recognition}
\author{\uppercase{Jia Le Ngwe}\authorrefmark{1},
\uppercase{Kian Ming Lim}\authorrefmark{1}, \IEEEmembership{Senior Member, IEEE},
\uppercase{Chin Poo Lee}\authorrefmark{1}, \IEEEmembership{Senior Member, IEEE},
\uppercase{Thian Song Ong}\authorrefmark{1}, \IEEEmembership{Senior Member, IEEE},
\uppercase{Ali Alqahtani}\authorrefmark{2,3}}

\address[1]{Faculty of Information Science and Technology, Multimedia University, Jalan Ayer Keroh Lama, 75450, Melaka, Malaysia}
\address[2]{Department of Computer Science, King Khalid University, Abha 61421, Saudi Arabia}
\address[3]{Center for Artificial Intelligence (CAI), King Khalid University, Abha 61421, Saudi Arabia}
\tfootnote{This research is supported by Telekom Malaysia Research \& Development under grant number RDTC/231084 and Deanship of Scientific Research, King Khalid University, Saudi Arabia, under Grant number RGP2/332/44.}

\markboth
{Author \headeretal: Preparation of Papers for IEEE TRANSACTIONS and JOURNALS}
{Author \headeretal: Preparation of Papers for IEEE TRANSACTIONS and JOURNALS}

\corresp{Corresponding author: Kian Ming Lim (e-mail: kmlim@mmu.edu.my).}

\begin{abstract}
Facial Expression Recognition (FER) is a machine learning problem that deals with recognizing human facial expressions. While existing work has achieved performance improvements in recent years, FER in the wild and under challenging conditions remains a challenge. In this paper, a lightweight patch and attention network based on MobileNetV1, referred to as PAtt-Lite, is proposed to improve FER performance under challenging conditions. A truncated ImageNet-pre-trained MobileNetV1 is utilized as the backbone feature extractor of the proposed method. In place of the truncated layers is a patch extraction block that is proposed for extracting significant local facial features to enhance the representation from MobileNetV1, especially under challenging conditions. An attention classifier is also proposed to improve the learning of these patched feature maps from the extremely lightweight feature extractor. The experimental results on public benchmark databases proved the effectiveness of the proposed method. PAtt-Lite achieved state-of-the-art results on CK+, RAF-DB, FER2013, FERPlus, and the challenging conditions subsets for RAF-DB and FERPlus.
\end{abstract}


\titlepgskip=-21pt

\maketitle

\section{Introduction}
\label{sec:introduction}
\PARstart{F}{acial} expression is a complex and fascinating aspect of nonverbal human communication that involves a range of facial muscle movements. These changes can convey a wide range of emotions and mental states, including happiness, sadness, anger, surprise, fear, and disgust. Given the importance of facial expressions in communication, it is not surprising that there has been a growing interest in automated facial expression recognition (FER) technology. FER has the potential to revolutionize a wide range of fields, from education to healthcare. For example, FER could be used in educational settings to measure the effectiveness and quality of teaching \cite{wu2021adaptive, li2021recognizing}, or in healthcare settings to assist in the analysis of the psychological condition of a patient \cite{jonitta2020deep, ye2023analysis}. Along with the advances made in GPU technology, the enormous potential for downstream applications of FER also contributed to its increasing popularity.

The main challenges that FER poses differently from other image classification tasks are the inter-class similarities and intra-class differences in human facial expressions. Inter-class similarities refer to the subtle differences between facial expressions, which makes it difficult to highlight the small differences between facial expressions and recognize them correctly. On the other hand, intra-class differences, also known as subject variability, refer to the characteristic of FER databases that images from an expression class are made up of different subjects with different facial structures, gender, age, and race. This variability can hinder the learning performance of a solution, as the model may struggle to generalize across different subjects, leading to reduced accuracy and reliability. For example, the differences between an angry face and a disgusted face may be minimal, whereas the differences between two different individuals within the same expression class can be quite significant.

In addition, existing work has exposed other FER challenges on in-the-wild databases, namely the recognition of negative expressions, FER under challenging conditions, and reliance on large neural networks. The scarcity of negative expression images on the Internet has made it difficult to collect a representative database that can reflect real-world scenarios. Therefore, it can result in a class imbalance in the in-the-wild FER databases, which can cause the recognition rate of negative expressions to be lower than that of positive expressions. FER under challenging conditions refers to the recognition of facial expressions when the subjects are posed at certain angles or when the subject faces are partially occluded by other objects. The accurate recognition of these samples is important, especially since the challenging conditions are likely conditions identical to the downstream applications. Meanwhile, in the pursuit of classification performance, existing work is also slowly leaning towards large neural networks to achieve these performance improvements. However, considering the computing resources of downstream applications, FER methods should be readily available for these applications without requiring powerful resources.

In this paper, PAtt-Lite, a lightweight patch and attention network is proposed to improve the FER performance under challenging conditions. First, a truncated MobileNetV1 is employed as the backbone model. A patch extraction block is proposed for the truncated backbone model to enforce the model to extract significant local facial features to classify facial expressions under challenging conditions accurately. It is designed to be lightweight while responsible for splitting the MobileNetV1 feature maps into 4 non-overlapping regions. A self-attention classifier is proposed for the backbone model to improve the learning of the output feature maps. With a dot product self-attention layer sandwiched between two fully connected layers, the attention classifier is able to learn the patched feature maps better than a vanilla classifier, hence enhancing the performance of the proposed PAtt-Lite under challenging conditions. Finally, to evaluate the performance of the proposed method, one lab-controlled database, i.e., CK+, and three in-the-wild databases, i.e., RAF-DB, FER2013, and FERPlus are employed as the benchmark databases of this research. Extensive experiments to determine the performance of the proposed method under challenging conditions such as occlusion and posed subjects using the challenging condition subsets introduced by \cite{wang2020region} are also conducted.

The main contributions of this work are as follows:
\begin{enumerate}
    \item A lightweight patch extraction block is proposed and added to the truncated MobileNetV1 to extract significant local facial features for accurately classifying facial expressions of occluded or posed subjects.
    \item Attention classifier is proposed to relate the global average pooled output feature maps and detect their underlying patterns for better classification performance.
    \item Extensive experimental results demonstrate the superiority of the proposed method over state-of-the-art methods on all benchmark databases, including the challenging condition subsets, despite its lightweight nature and significantly lesser parameters than state-of-the-art methods.
\end{enumerate}

The remaining of this paper is organized as follows. Section 2 reviews related work with a focus on the application of the patch extraction block and the attention mechanism. Section 3 provides an overview of the architecture of the proposed PAtt-Lite, followed by detailed explanations of each module in the proposed solutions. Section 4 introduces the benchmark databases and details the experimental setting of the proposed method, along with an ablation analysis of the modules presented in the solutions, and a comparison of the proposed method with the state-of-the-art. Finally, the conclusion for this paper is included in Section 5.

\section{Related Work}
\subsection{Convolutional Neural Network}
Convolutional Neural Network (CNN) is a class of deep learning models designed to process grid-like data such as images. It employs convolutional layers to automatically detect patterns or features through spatial hierarchies, enabling the network to learn progressively complex information, and pooling layers to reduce dimensionality and computational complexity while maintaining important features. These networks have achieved significant results in computer vision tasks such as image classification, image segmentation, and object detection. The main advantage of CNNs is their ability to learn complex features automatically without the need for manual feature engineering. Besides, CNNs are also highly adaptable to different input sizes while being able to handle complex patterns and data variations. 

With the advancement in GPU technology and the availability of mature deep learning libraries, existing work for FER has focused more on deep learning solutions recently. These solutions often outperform the handcrafted methods, especially in in-the-wild databases. Most of the CNN-based methods, such as \cite{xie2018facial, zhao2018feature, li2018occlusion, wang2020region, gera2021landmark, gera2021imponderous, ding2020occlusion}, attempt to improve the FER performance by exploiting local information in different ways with their additional modules. 

The development of CNN architectures has brought forward many innovations, including residual connections\cite{he2016deep}, bottleneck design\cite{szegedy2015going, he2016deep}, batch normalization\cite{ioffe2015batch} and its alternatives\cite{ba2016layer, wu2018group}, depthwise separable convolutions\cite{howard2017mobilenets}, and many more. However, the architectures that integrate some of these innovations are complex and have a higher number of training parameters, which in turn require larger databases and longer training time. By contrast, the architecture of MobileNetV1\cite{howard2017mobilenets} is simpler and lighter than most of the well-known CNN architectures by comparison. Thus, the ImageNet-pre-trained MobileNetV1 is selected as the baseline architecture for this research, due to its high performance despite its lightweight and simple architecture. This simple architecture also has provided an easy finetuning process since overfitting and underfitting on the benchmark databases are easy to control with this architecture. 

The base feature extractor is paired with the proposed patch extraction block as our attempt to improve FER performance under challenging conditions. The patch extraction block is designed to solely extract local features. This distinguishes it from the methods used in \cite{xie2018facial, li2018occlusion, gera2021imponderous}, which employed patch attention mechanisms. Specifically, the proposed patch extraction is inspired by that of the vision transformer architecture but remains different from its inspiration in terms of implementation details, which will be explained in the next subsection. 

\subsection{Vision Transformers}
In recent years, the Transformer architecture has gained increasing attention in various deep learning tasks, particularly natural language processing. It is a type of neural network introduced by \cite{vaswani2017attention} that was designed for sequence-to-sequence tasks. The architecture consists of a stacked encoder and/or decoder layers that allow for efficient and scalable processing of large input data while learning even more complex patterns in the data. The vision transformer (ViT) architecture introduced by \cite{dosovitskiy2020image} is a novel approach that adapts the Transformer architecture to computer vision tasks. It divides images into smaller, non-overlapping patches and reshapes them into 1D sequences before processing them as a sequence using a Transformer model. The success of this architecture has also attracted researchers’ attention for the development of ViT alternatives such as DeiT\cite{touvron2021training} and Swin Transformer\cite{liu2021swin, liu2022swin}. Overall, ViT and its alternatives have achieved state-of-the-art performance on various tasks, including image classification, thus demonstrating the versatility and effectiveness of the Transformer architecture beyond natural language processing. 

Hence, researchers also have begun to introduce the Transformer or the ViT architecture for FER in recent years \cite{aouayeb2021learning, xue2021transfer, zheng2022poster, xue2022vision, mao2023poster, ma2021facial, li2021mvt}, motivated by their performance achieved across different tasks. Based on the results posted in existing work, the application of vision transformers in FER is proven to be useful with ViT+SE\cite{aouayeb2021learning} posting the state-of-the-art performance of 99.80\% mean accuracy across 10 folds on the CK+ database, POSTER++\cite{mao2023poster} being the best-performing method on the RAF-DB database with 92.21\% accuracy, and POSTER\cite{zheng2022poster} being the second best-performing method on the FERPlus database by achieving 91.62\% accuracy. However, this performance often comes with large architectures with significantly more parameters than CNN-based methods. Nevertheless, the raw performance of the ViT architecture also attracted our attention to draw some inspiration for integrating into the MobileNetV1 backbone for better FER performance. 

Although the patch extraction block is ultimately inspired by ViT, there exist some differences in terms of the implementation details. The first difference is the design and placement of the patch extraction block. The patch extraction mechanism in ViT is a single-layer convolution that is placed at the beginning of the architecture, whereas the patch extraction block in the proposed PAtt-Lite is a multi-layer convolution that is placed within the architecture. This placement allows the proposed method to fully utilize the pretrained weights of the backbone MobileNetV1, which were trained on ImageNet samples of size $224\times224$. Secondly, ViT splits the input image of size $224\times224$ into 196 non-overlapping patches of size $16\times16$, whereas PAtt-Lite splits the output feature maps from MobileNetV1 of size $14\times14$ into 4 non-overlapping patches of size $1\times1$. The larger receptive regions of the proposed patch extraction block also help the proposed PAtt-Lite in extracting significant and high-level facial features. 

The attention mechanism is intended to model the human attention mechanism, by highlighting parts of the input feature while ignoring the others, which enables better learning of the correlation between two input sequences. There are many variations of the attention mechanisms which sport different score functions, like additive attention \cite{bahdanau2014neural} and dot-product attention \cite{luong2015effective}. The key component of the Transformer architecture is its self-attention, which is an attention mechanism that relates different positions of the same sequence. In this paper, an attention classifier inspired by the Transformer architecture is proposed to further improve the learning of output feature maps from the modified lightweight feature extractor. Specifically, the proposed attention classifier attempts to replicate the performance of vision transformers without requiring its series of self-attention blocks. Instead, a dot-product self-attention operation is integrated between the fully connected layers of the classifier. Through this design decision, the model can be kept lightweight while retaining high feature extractive and classification performance for FER. 

\section{Methodology}
\begin{figure*}
    \centering
    \includegraphics[width=\textwidth]{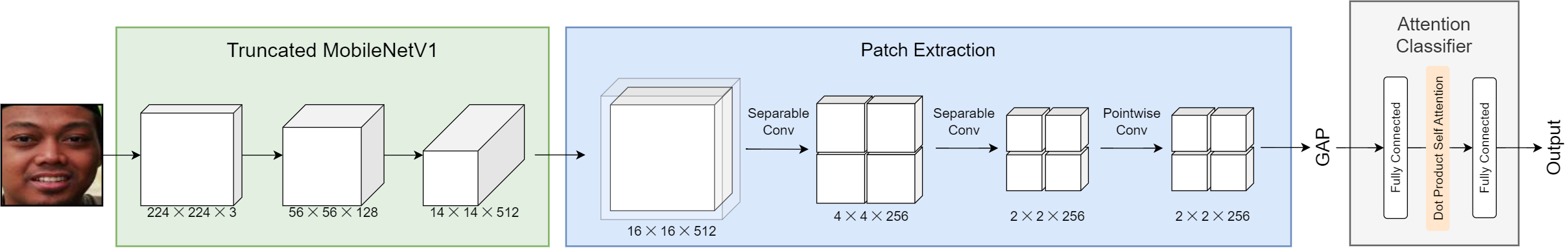}
    \caption{Architecture of the proposed PAtt-Lite. The image sample will first go through the truncated MobileNetV1 for feature extraction, in which the output feature maps will be padded and used as input for the proposed patch extraction block. The output feature maps of dimensions $2 \times 2 \times 256$ from the patch extraction block will then be global average pooled before being taken by the attention classifier. }
    \label{fig:architecture}
\end{figure*}

\subsection{Overview}
Fig. \ref{fig:architecture} illustrates the overall architecture of the proposed PAtt-Lite. The proposed PAtt-Lite is built upon a truncated pre-trained MobileNetV1, combined with the proposed patch extraction block and attention classifier. Specifically, layers after the depthwise convolution of block 9 are truncated. The proposed patch extraction block and attention classifier are added to the truncated backbone model. 

Given an image sample, the input will first go into the truncated MobileNetV1 to leverage the feature-extracting capability of the pre-trained model on the lower-level details of the image. The output feature maps are then used as input for our patch extraction block, where meaningful local features are extracted. The output feature maps from the patch extraction block are in the dimensions of $2 \times 2 \times D$, where $D$ represents the depth of the feature maps. The attention classifier takes the feature maps that have been global average pooled as input, and outputs the probabilities of the facial expressions. 

\subsection{MobileNetV1}
CNNs have been used in various computer vision tasks, such as object detection, image classification, and semantic segmentation, with state-of-the-art performance. However, CNNs can be computationally intensive and require large memory footprints, which makes them impractical for deployment on mobile or edge devices. MobileNetV1 \cite{howard2017mobilenets} is a family of lightweight CNN architectures designed to be used on mobile and embedded devices. By leveraging depthwise separable convolutions, MobileNetV1 achieves a significant reduction in the number of model parameters and the number of multiplication and addition operations required for inference (Mult-Adds). 

Depthwise separable convolution is a departure from the standard convolutional operation, as they split a standard convolution into two separate operations by performing depthwise convolutions followed by pointwise convolutions. Depthwise convolution is different from conventional convolution in that depthwise convolution applies a single convolutional filter for each input channel, whereas conventional convolution has filters that are as deep as its input. Meanwhile, pointwise convolution can be achieved using the standard convolutional operation by setting the kernel size to 1. Effectively, pointwise convolutions enable the mixing of input channels as conventional convolutions do. 

The architecture of MobileNetV1 is simpler than most of the well-known CNN architectures by comparison. Other than being a lightweight architecture, the absence of complex designs like residual connections and bottleneck layers also contributed to the easiness of finetuning the pre-trained weights on our benchmark databases optimally. Mathematically, the feature extractive process from the truncated MobileNetV1 is formulated as follows: 
\begin{equation} \label{eq:mobilenet}
    X_{FE} = \text{MobileNetV1}(X)
\end{equation}
where $X$ is the original sample image and $X_{FE}$ is the output feature maps from the backbone feature extractor. 

\subsection{Patch Extraction}
The key advantage of using a pre-trained CNN for transfer learning is that earlier layers have learned generic features of the training samples, such as the edges, whereas the later layers have learned specific features of training samples. In the context of PAtt-Lite, layers after the depthwise convolution of block 9 are skipped. The patch extraction block is added to better adapt to the FER databases than simply fine-tuning the final layers. This modification to the feature extractor also results in a shorter training period, as a higher learning rate can be used as opposed to when the pre-trained weights are being finetuned. 

Our proposed patch extraction block consists of three different convolutional layers, the first two being depthwise separable convolutional layers and the last being a pointwise convolutional layer. Operating on feature maps from the MobileNetV1 that are padded to the dimension of $16\times16$, the first separable convolutional layer is responsible for splitting the feature maps into four patches while learning higher-level features from its input. Subsequently, the second separable convolutional layer and the pointwise convolutional layer are responsible for learning the higher-level features from the patched feature maps, resulting in output with a dimension of $2\times2$. Instead of the standard convolutional layer used in conventional CNNs, the depthwise separable convolutional layer is selected for PAtt-Lite. This design decision improves the classification performance of the proposed method on challenging subsets while reducing the number of model parameters. 

The design process of the patch extraction block started with a grid search for the optimum MobileNetV1 output layer with the number of patches for the patch extraction block. The baseline block consisted of a convolutional layer and a pointwise convolutional layer, which is retained in the final design. Our grid search has experimented with five convolutional kernel sizes, which are 3, 4, 5, 7, 8, and all layers of MobileNetV1 with the feature map size of $2 \times 2 \times D$. The summary of these experiments is included in Table~\ref{tab:patch-ablation}. 

The baseline patch extraction block is then redesigned based on the optimum patch size and output layer. With the pointwise convolutional layer kept as is, the first convolutional layer was swapped out for a separable convolutional layer, which has a significantly smaller number of parameters while still providing the same spatial and channel convolution. Meanwhile, another separable convolutional layer was added to the patch extraction block to keep the number of patches constant while removing the need for a larger convolutional kernel. Thus, we arrived at our final design for the patch extraction block. 

\subsection{Global Average Pooling}
Global average pooling (GAP) is a technique that was first introduced in \cite{lin2013network} to address the problem of overfitting in CNNs. GAP is a type of pooling operation that computes a single value for each feature map by taking the average of all the values in that map. Unlike conventional pooling operations, which reduce the spatial resolution of feature maps, GAP is normally applied at the end of a CNN architecture. The application of GAP can result in a much smaller output volume, with the output value also acting as a confidence map for each category that CNN is trained to recognize. 

GAP in the proposed PAtt-Lite is responsible for averaging the patch representation from our patch extraction block, which removes the need of flattening the feature maps and feeding them into fully connected layers, hence resulting in a slight reduction in the number of parameters while further minimizing the possibility of overfitting. 

Let $X_{PE}$ be the output feature maps from the patch extraction block and ${\bar{X}}_{PE}$ be the output from the GAP operation, this operation can be represented with the equation as follows: 
\begin{equation} \label{eq:gap}
    {\bar{X}}_{PE} = \text{GAP}(X_{PE})
\end{equation}

\subsection{Attention Classifier}
An attention classifier is introduced in the proposed method for better learning of representation from the backbone MobileNetV1 and the patch extraction block. The attention classifier comprises a dot-product\cite{luong2015effective} self-attention\cite{cheng2016long} layer placed between two fully connected layers of the newly added classifier. 

Dot product attention is a specific type of self-attention mechanism where the attention weights are computed as a dot product between the query vector and the key vector, divided by the square root of the dimension of the key vectors. Self-attention, also known as intra-attention in \cite{cheng2016long}, is a mechanism that allows a neural network to focus on specific parts of its input during computation selectively. The idea behind self-attention is to allow the network to learn a set of attention weights that indicate how important each input element is to the output of the network. It has become a popular technique in natural language processing and computer vision tasks as it can help improve performance by selectively attending to the most relevant parts of the input.

Let $Q$, $K$, and $V$ be the query, key, and value vectors, respectively, and $d_q = d_k$. The dot-product self-attention score can be computed as follows:
\begin{equation} \label{eq:attn}
    \text{Attention}(Q,K,V) = \text{softmax}(\frac{QK^T}{\sqrt{d_q}}) V
\end{equation}
where $d_k$ is the dimensionality of the key vectors. The softmax function is applied to the dot-product similarity scores to obtain a set of attention weights that sum up to 1. These weights are used to compute a weighted sum of the value vectors, resulting in the final attention output. Hence, together with the fully connected layers, the attention classifier can be represented with the following equations: 
\begin{equation} \label{eq:relu}
    X_R = \text{ReLU}({\bar{X}}_{PE})
\end{equation}
Let $Q$, $K$, and $V$ be the query, key, and value vectors computed from the input vector $X_R$, the final attention output, $X_A$ can be computed as follows: 
\begin{equation} \label{eq:attn-value}
    X_A = \text{Attention}(Q,K,V)
\end{equation}
\begin{equation} \label{eq:softmax}
    Y = \text{softmax}(X_A)
\end{equation}
where ${\bar{X}}_{PE}$ is the output values from GAP, $X_R$ is the output values from the first fully connected layer with ReLU activation function, and $Y$ represents the predicted target label as output from the final fully connected layer with softmax activation function. 

\section{Experiments and Comparison}
\subsection{Databases}
Both laboratory-controlled and in-the-wild databases are used to evaluate the proposed PAtt-Lite, namely CK+, RAF-DB, FER2013, and FERPlus. Summaries of the class data distribution for in-the-wild databases and their challenging subsets are presented in Table~\ref{tab:class-dist-wild} and Table~\ref{tab:class-dist-challenge}. A summary of the training distribution for in-the-wild databases is also shown in Table~\ref{tab:train-dist-wild}. 

\textbf{CK+}\cite{lucey2010extended} is a well-known laboratory-controlled database extended from the CK database. The database consists of 593 image sequences from 123 subjects, of which 327 are labeled with one of the 7 discrete emotions: Anger, Disgust, Fear, Happy, Sadness, Surprise, and Contempt, with the first images in the sequence being the neutral expression. This research evaluates the 7 emotions CK+ with 10-fold subject-independent cross-validation to have a fair comparison with most existing work. 

\textbf{RAF-DB}\cite{li2017reliable} is another widely used database in recent years. The database contains great variation in terms of gender, age, race, and pose of the subjects. Nearly 30,000 sample images are included in the database with crowdsourced annotations from 40 taggers. This research evaluates the basic expression subset of the database, which contains 12,271 training images and 3,068 testing images. The challenging condition test subsets of the RAF-DB database introduced by \cite{wang2020region} are also evaluated in this research. 

\textbf{FER2013}\cite{goodfellow2013challenges} is introduced during the FER challenge hosted on Kaggle. It is a database collected through the Google image search API, with nearly 36,000 sample images included. The sample images are annotated with 7 basic expression labels, i.e., Angry, Disgust, Fear, Happy, Neutral, Sad, and Surprise, by 1 tagger. Compared to RAF-DB, this is a relatively more challenging database, as some of the samples are incorrectly labeled and some are without a face. 

\textbf{FERPlus}\cite{barsoum2016training} is extended from FER2013 by relabeling the original database through crowdsourcing from 10 taggers. The sample images are annotated with 8 basic expression labels, through the addition of the Contempt label. This process has corrected incorrectly labelled samples and has removed faceless samples, resulting in 35,710 sample images in the database. The challenging condition test subsets of the FERPlus database introduced by \cite{wang2020region} are also evaluated in this research. 

\begin{table*}
    \caption{Summary of class data distribution for in-the-wild databases. }
    \label{tab:class-dist-wild}
    \centering
    \begin{tabular}{c c c c c c c c c}
        \hline
        Databases & Anger & Disgust & Fear & Happiness & Neutral & Sadness & Surprise & Contempt \\
        \hline
        RAF-DB & 867 & 877 & 355 & 5957 & 3204 & 2460 & 1619 & - \\
        FER2013 & 4953 & 547 & 5121 & 8989 & 6198 & 6077 & 4002 & - \\
        FERPlus & 3123 & 253 & 825 & 9367 & 13014 & 4414 & 4493 & 221 \\
        \hline
    \end{tabular}
\end{table*}

\begin{table*}
    \caption{Summary of class data distribution for challenging subsets of RAF-DB and FERPlus.}
    \label{tab:class-dist-challenge}
    \centering
    \begin{tabular}{c c c c c c c c c}
        \hline
        Subsets & Anger & Disgust & Fear & Happiness & Neutral & Sadness & Surprise & Contempt\\
        \hline
        \multicolumn{9}{c}{RAF-DB}\\
        \hline
        Occlusion & 31 & 47 & 35 & 236 & 191 & 122 & 72 & - \\
        Pose 30 & 75 & 78 & 38 & 446 & 282 & 179 & 149 & - \\
        Pose 45 & 37 & 36 & 20 & 164 & 129 & 90 & 82 & - \\
        \hline
        \multicolumn{9}{c}{FERPlus}\\
        \hline
        Occlusion & 23 & 3 & 33 & 122 & 162 & 125 & 135 & 2 \\
        Pose 30 & 102 & 5 & 28 & 286 & 466 & 151 & 127 & 5 \\
        Pose 45 & 52 & 1 & 16 & 141 & 271 & 89 & 59 & 4 \\
        \hline
    \end{tabular}
\end{table*}

\begin{table}
    \caption{Summary of training distribution for in-the-wild databases.} 
    \label{tab:train-dist-wild}
    \centering
    \begin{tabular}{c c c c}
        \hline
        Databases & Training & Validation & Testing \\
        \hline
        RAF-DB & 12271 & - & 3068 \\
        FER2013 & 28659 & 3584 & 3582 \\
        FERPlus & 28558 & 3579 & 3573 \\
        \hline
    \end{tabular}
\end{table}

\subsection{Implementation Details}
The proposed method is implemented with the TensorFlow library on an NVIDIA TESLA P100 GPU from the Kaggle platform. A resizing operation is added to ensure that all sample images are resized to $224\times224$. Random horizontal flip and random contrast are performed for data augmentation. 

A two-stage training-finetuning process from \cite{transferlearning} is employed for the training process. The pre-trained weights are frozen to solely adapt the new components to the output from MobileNetV1 during the training stage. For the finetuning process, several layers were unfrozen for finetuning the feature extractor to the benchmark databases. To extract the best feature extractive performance from the backbone MobileNetV1, 40 layers, 59 layers, 46 layers, and 49 layers were unfrozen for CK+, RAF-DB, FER2013, and FERPlus, respectively. 

We use sparse categorical cross entropy as the loss function, Adam as the optimizer, and a batch size of 8 for all experiments. For better stability of the proposed method, global gradient norm clipping is also added to the experiments. The initial learning rate is set to $1\times{10}^{-3}$ for the initial training process. The learning rate is scheduled by decreasing it when the model accuracy is not improving for longer than the number of epochs that were set as patience. For the finetuning process, the learning rate is scheduled based on the inverse time decay schedule with the initial learning rate of $1\times{10}^{-5}$. The number of epochs for both the initial training and finetuning process is determined by the early stopping callback with restoration to the best weights when the training process is terminated. 

\subsection{Ablation Study}
For the ablation study, the effectiveness of the patch extraction block and the attention classifier are evaluated by comparing them to the baseline performance of the finetuned MobileNetV1. Experiments are conducted on all benchmark databases for a proper evaluation on the effect of the proposed modules. The summary of the experimental results for our grid search on the optimum output layer and number of patches is also included in Table~\ref{tab:patch-ablation}. Additionally, experimental results for the comparison between patch extraction and patch attention on in-the-wild databases are also included to justify the proposed patch extraction instead of conventional patch attention. The experimental results for this study are presented in Table~\ref{tab:ablation-study} and Table~\ref{tab:attention-extraction}. 

\begin{table*}
    \caption{Ablation study for the proposed method on all benchmark databases. The best result is highlighted in bold.}
    \label{tab:ablation-study}
    \centering
    \begin{tabular}{c c c c c c}
        \hline
        Attention Classifier & Patch Extraction & CK+ & RAF-DB & FER2013 & FERPlus \\
        \hline
        \xmark & \xmark & 99.90 & 85.17 & 68.54 & 82.88 \\
        \xmark & \cmark & 100.00 & 81.10 & 61.52 & 77.72 \\
        \cmark & \xmark & 100.00 & 91.00 & 83.00 & 90.71 \\
        \cmark & \cmark & \textbf{100.00} & \textbf{95.05} & \textbf{92.50} & \textbf{95.55} \\
        \hline
    \end{tabular}
\end{table*}

\subsubsection{Effectiveness of proposed modules}
As shown in Table~\ref{tab:ablation-study}, the proposed patch extraction block is observed to have a slight decrease in performance compared to the MobileNetV1 baseline in in-the-wild databases. The performance drops in in-the-wild databases are mainly due to the difference in the number of trainable parameters between the original layers and the layers from the patch extraction block, which the newly initialized layers also resulted in the absence of pre-trained weights from the final layers. On the other hand, the MobileNetV1 baseline struggled to get a 100.00\% mean accuracy on CK+ based on our experiments. However, with the small scale of CK+, this performance is immediately achievable with the addition of the patch extraction block, hence validating its effectiveness. 

For the effectiveness of the proposed attention classifier, it improved the classification accuracy compared to the baseline. Moreover, the attention classifier has also significantly improved the classification accuracy on in-the-wild databases, achieving near-state-of-the-art performance on all in-the-wild databases that we benchmarked on. Specifically, the attention classifier provided an improvement of 5.83\% for RAF-DB, 14.46\% for FER2013, and 7.83\% for FERPlus. 

While the results show that the performance dropped with the patch extraction block alone, further performance improvement is achieved with both modules combined. This improvement is believed to stem from the self-attention layer between the fully connected layers, enabling the classifier to better adapt to the representations from the patch extraction block. The performance of the attention classifier is further boosted with the introduction of the newly initialized patch extraction block, which allowed these two modules to be trained at a higher learning rate, as opposed to the small learning rate normally used in finetuning over the pre-trained weights. Overall, the proposed patch extraction block provides a further improvement of 3.92\% accuracy on RAF-DB, 2.37\% on FER2013, and 4.84\% on FERPlus over the MobileNetV1 with attention classifier. 

\subsubsection{Patch Extraction Block}
Based on the experimental results in Table~\ref{tab:patch-ablation}, output feature maps from activated depthwise convolutional layers are the optimum layers for our patch extraction block, whereby patch sizes of 7 and 8 (with padding around the feature maps) generally yield the best results. Hence, this means splitting the feature maps into four patches is more optimal for our implementations. 

\begin{table}
    \caption{Summary of classification accuracy for different output layers and different kernel sizes on RAF-DB. The best accuracy is highlighted in bold.}
    \label{tab:patch-ablation}
    \centering
    \begin{tabular}{c c c}
        \hline
        Layer & Kernel & Best Accuracy (\%) \\
        \hline
        17 & NP7 & 90.03 \\
        17 & 0P3 & 89.96 \\
        17 & 0P5 & 89.80 \\
        17 & P4 & 89.11 \\
        17 & P8 & 88.92 \\
        \hline
        23 & NP7 & 93.71 \\
        23 & P8 & 93.45 \\
        23 & 0P5 & 91.69 \\
        23 & 0P3 & 91.17 \\
        23 & P4 & 89.93 \\
        \hline
        \textbf{29} & \textbf{NP7} & \textbf{94.78} \\
        29 & 0P3 & 93.84 \\
        29 & P8 & 93.64 \\
        29 & P4 & 89.41 \\
        29 & 0P5 & 80.22 \\
        \hline
        35 & P8 & 94.56 \\
        35 & NP7 & 93.64 \\
        35 & P4 & 90.51 \\
        35 & 0P5 & 85.43 \\
        35 & 0P3 & 82.95 \\
        \hline
        41 & P8 & 94.39 \\
        41 & NP7 & 93.09 \\
        41 & 0P5 & 92.86 \\
        41 & P4 & 89.50 \\
        41 & 0P3 & 82.66 \\
        \hline
        47 & P8 & 87.84 \\
        47 & NP7 & 86.38 \\
        47 & 0P5 & 84.03 \\
        47 & 0P3 & 79.04 \\
        47 & P4 & 78.13 \\
        \hline
    \end{tabular}
\end{table}

However, as the kernel size for the convolution operation gets bigger, so does its number of parameters, which contradicts with our idea of designing a small and lightweight model for facial expression recognition. Hence, we derived the design of the patch extraction block by experimenting with the replacement of the convolutional layer with a large kernel for the depthwise separable convolutional layer with a smaller kernel. In theory, this should still retain the performance of the original design as the separable convolutional layer is still performing the spatial and channel convolution that a conventional convolutional layer has. We also intend to keep the number of patches constant while trying to use a smaller kernel. Thus, the padding to the output feature maps from MobileNetV1 was kept, and a patch size of 4 was used for the first separable convolutional layer, and a patch size of 2 was used for the second separable convolutional layer.  From our experimental results, the replacement proved to be successful with the tradeoff of a minor performance drop for a significant reduction in the number of parameters. 

\subsubsection{Comparison between patch extraction and patch attention} 
\begin{table}
    \caption{Comparison between patch extraction and patch attention on in-the-wild databases. The best result is highlighted in bold.}
    \label{tab:attention-extraction}
    \centering
    \begin{tabular}{c c c c}
        \hline
          & RAF-DB & FER2013 & FERPlus \\
        \hline
        Patch Attention & 94.17 & 89.86 & 92.72 \\
        \textbf{Patch Extraction} & \textbf{95.05} & \textbf{92.50} & \textbf{95.55} \\
        \hline
    \end{tabular}
\end{table}

\begin{figure*}[!t]
\centering
    \subfloat[Occlusion RAF-DB]{\includegraphics[width=2.25in]{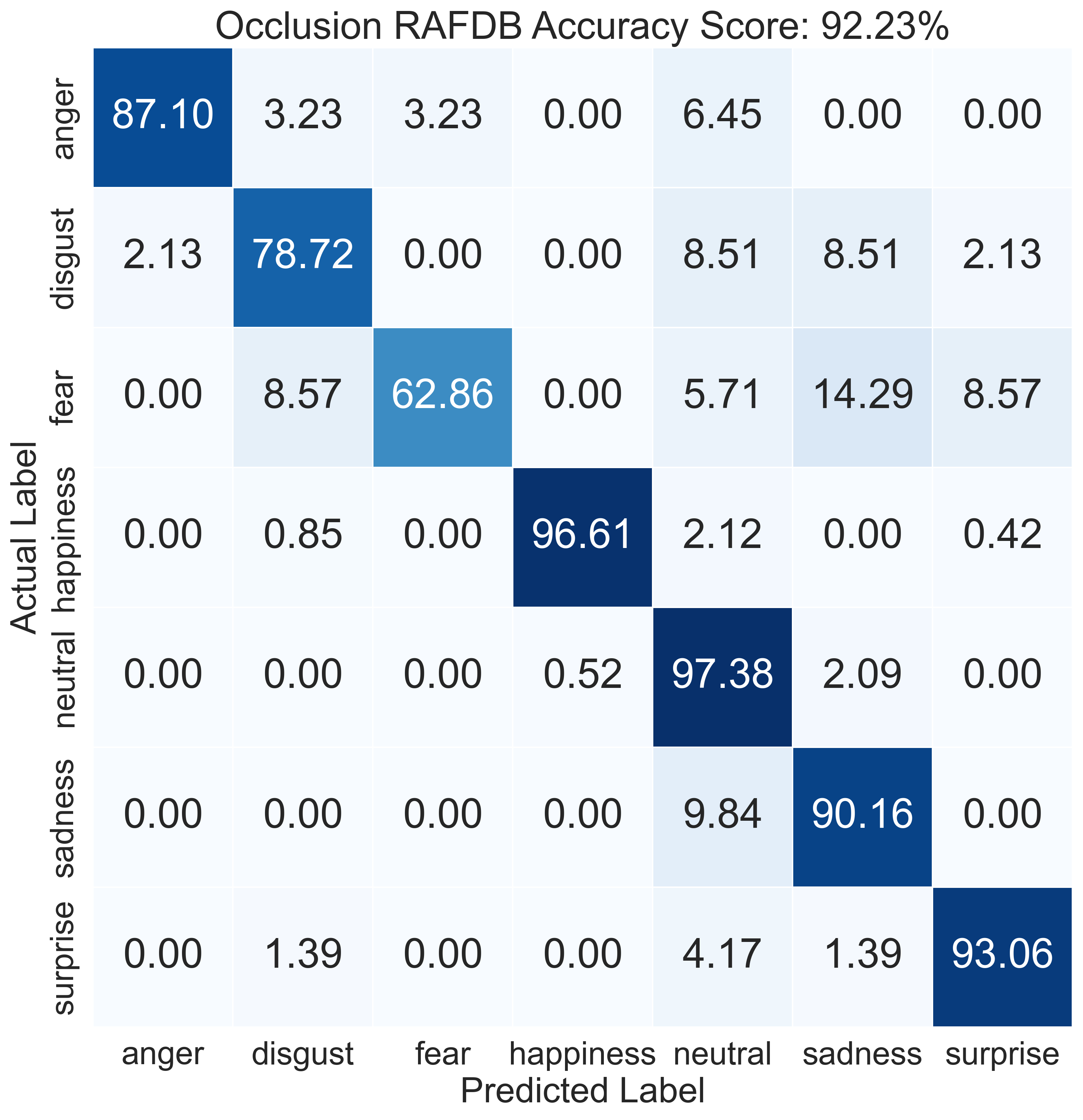}} \hfil
    \subfloat[Pose 30 RAF-DB]{\includegraphics[width=2.25in]{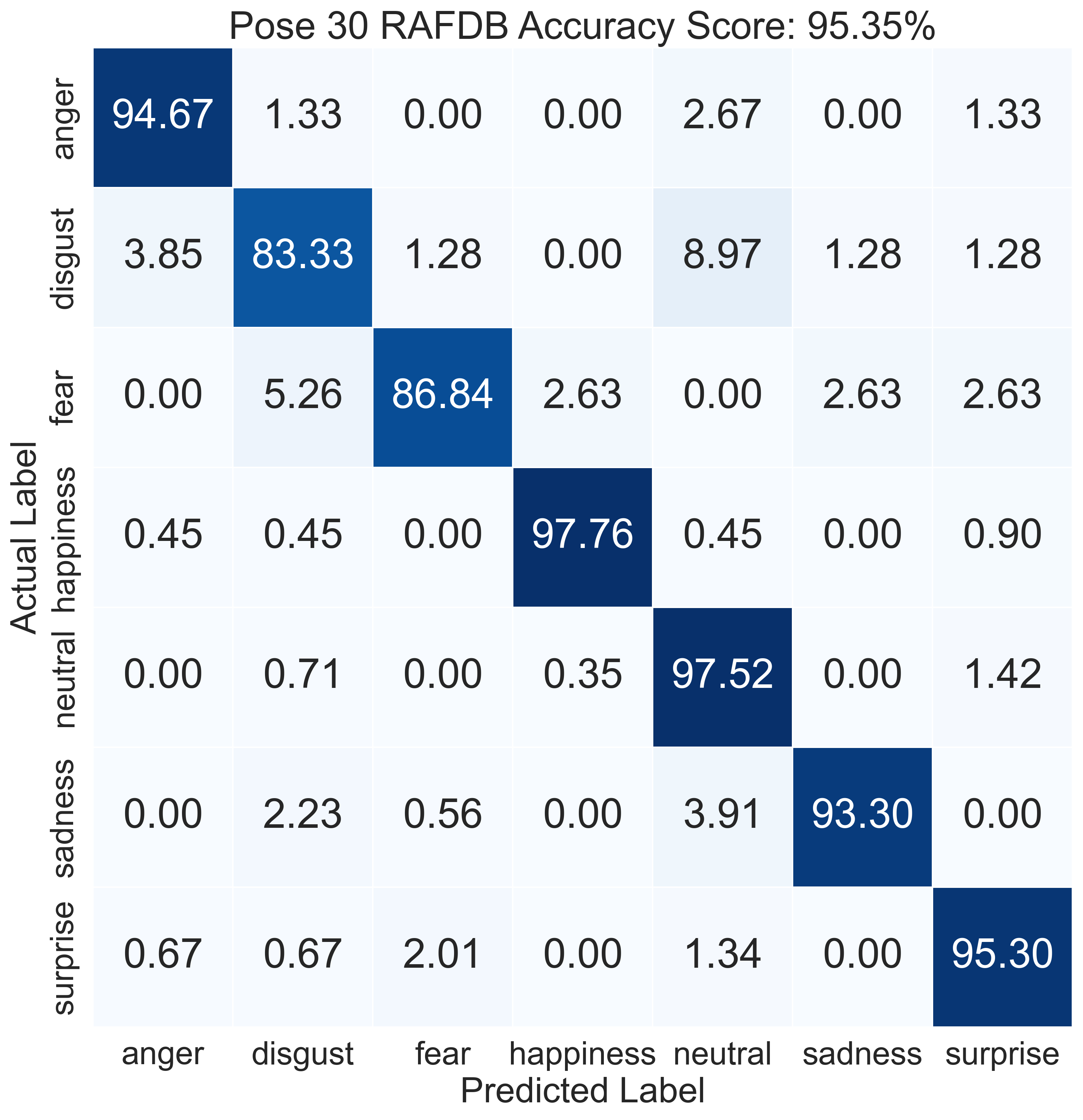}} \hfil
    \subfloat[Pose 45 RAF-DB]{\includegraphics[width=2.25in]{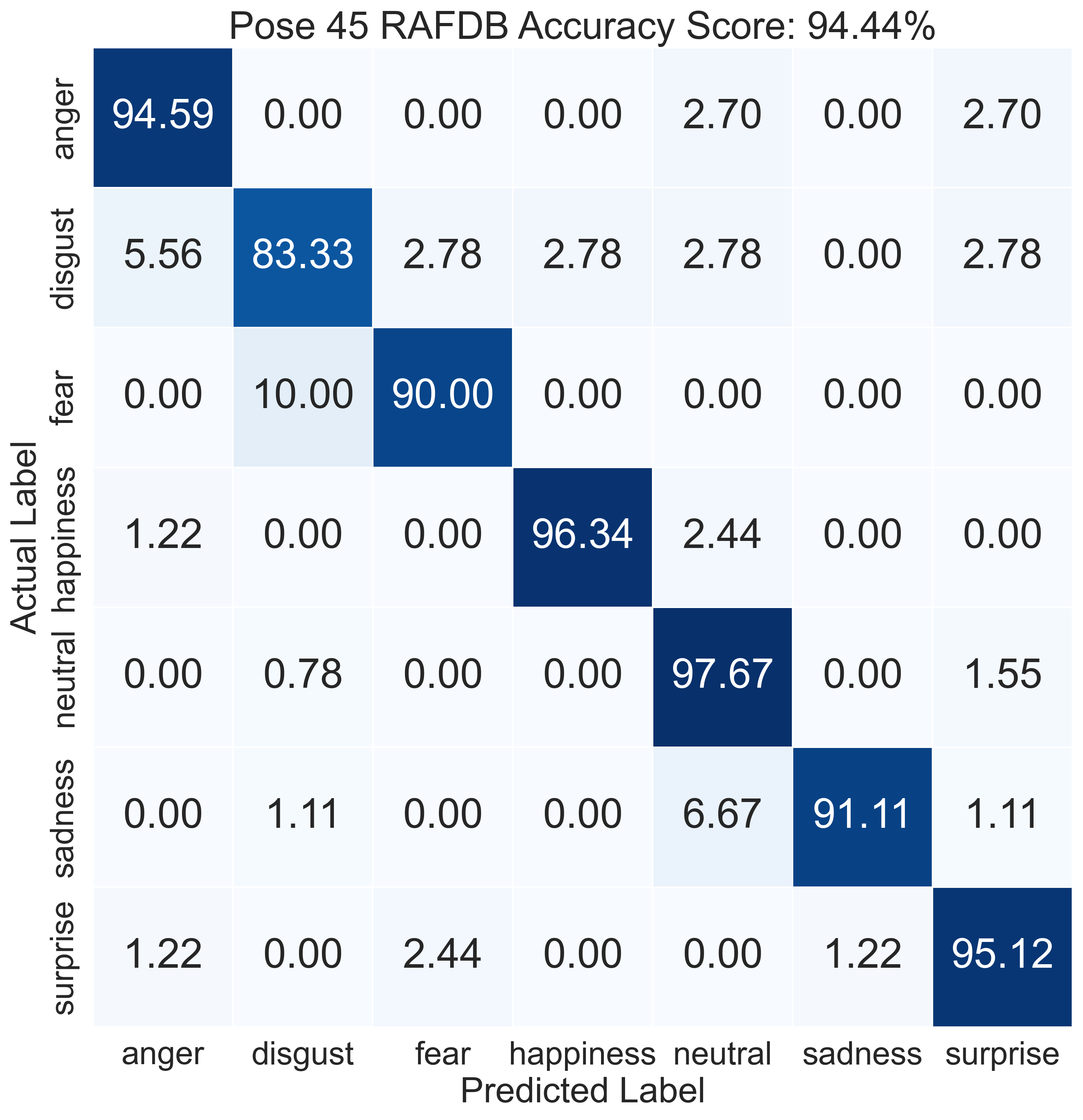}}
\caption{Confusion matrices of patch extraction on challenging subsets of RAF-DB.}
\label{fig:extraction-rafdb}
\end{figure*}

\begin{figure*}[!t]
\centering
    \subfloat[Occlusion RAF-DB]{\includegraphics[width=2.25in]{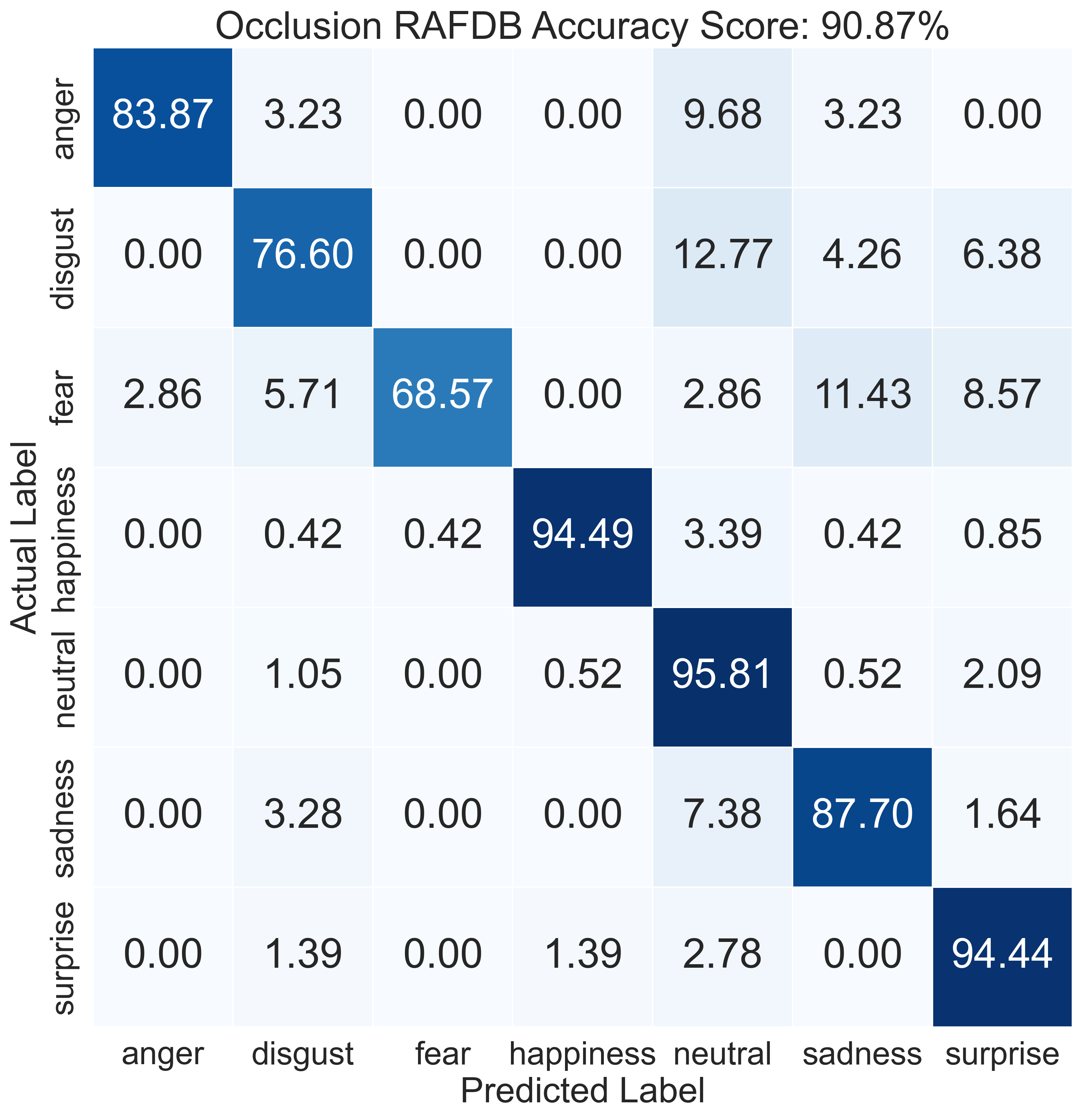}} \hfil
    \subfloat[Pose 30 RAF-DB]{\includegraphics[width=2.25in]{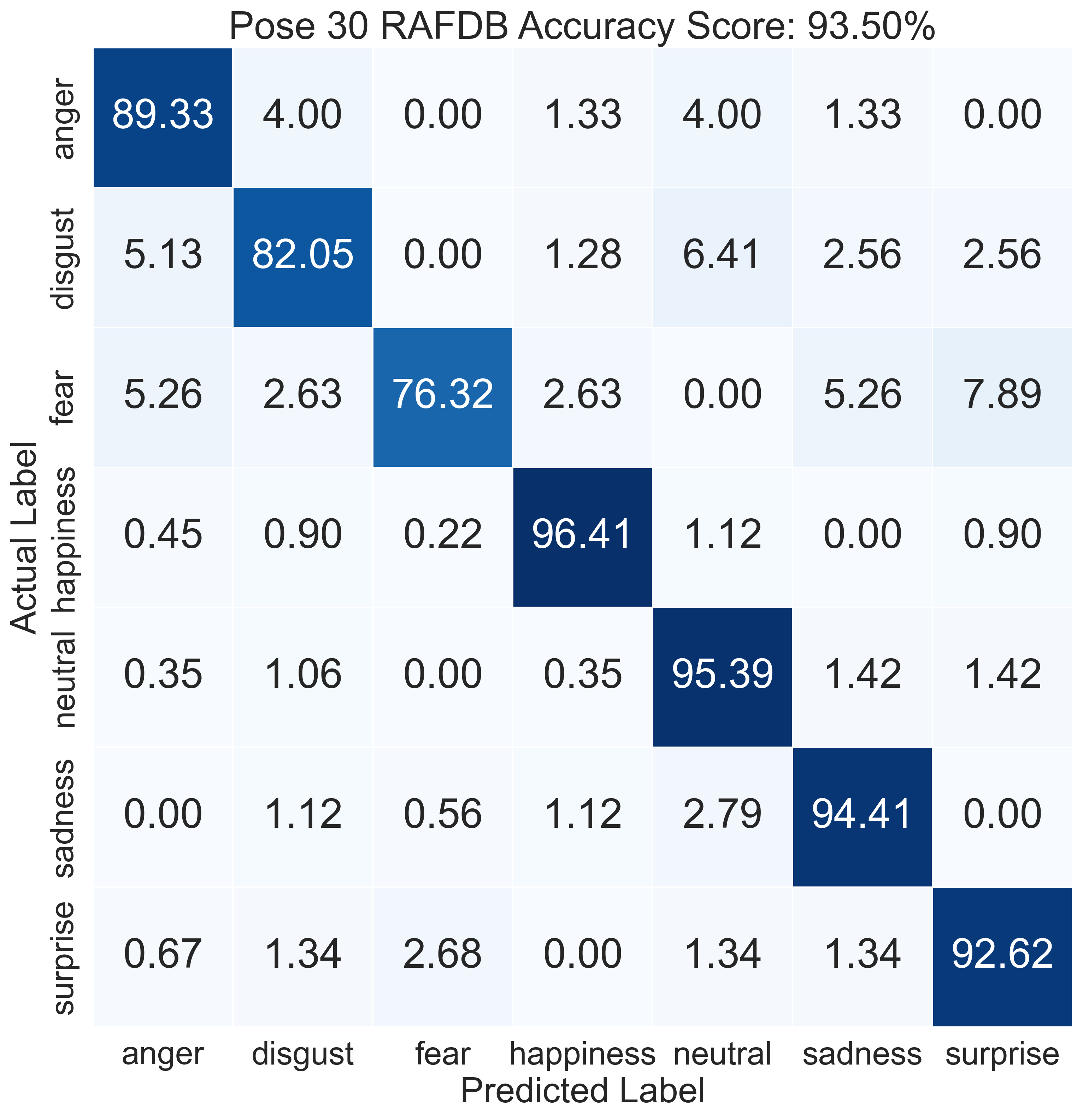}} \hfil
    \subfloat[Pose 45 RAF-DB]{\includegraphics[width=2.25in]{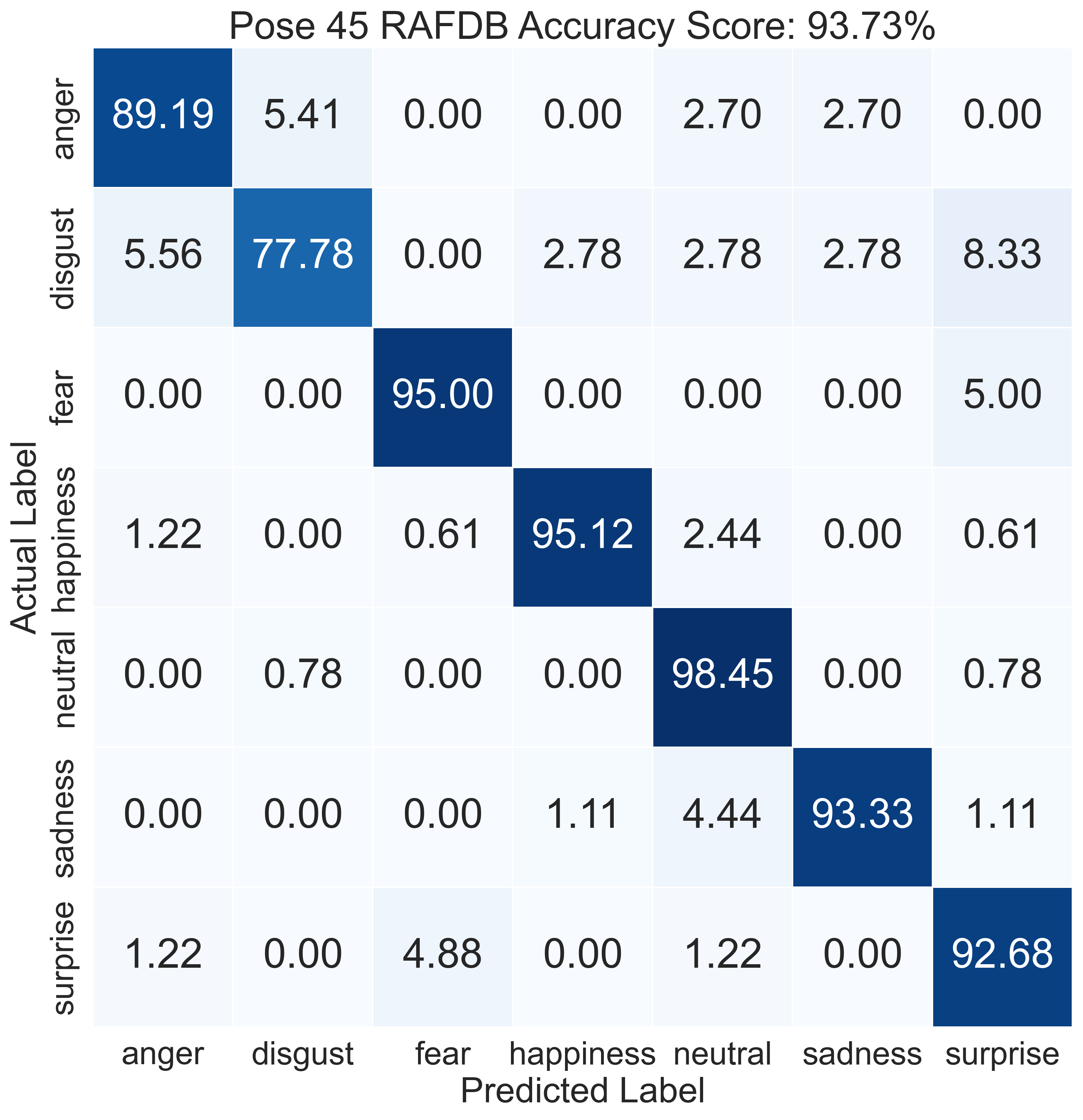}}
\caption{Confusion matrices of patch attention on challenging subsets of RAF-DB.}
\label{fig:attention-rafdb}
\end{figure*}

\begin{figure*}[!t]
\centering
    \subfloat[Occlusion FERPlus]{\includegraphics[width=2.25in]{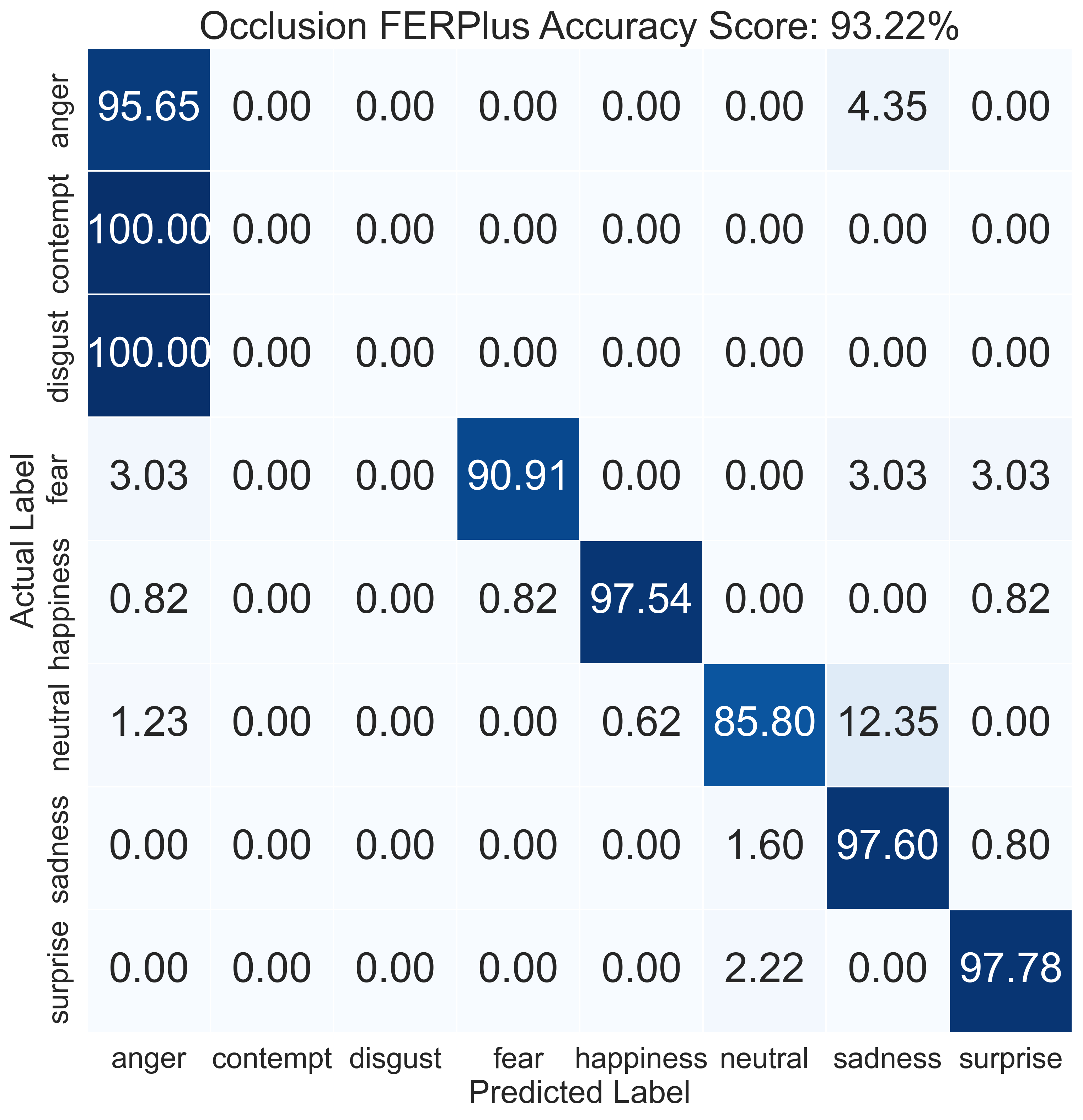}} \hfil
    \subfloat[Pose 30 FERPlus]{\includegraphics[width=2.25in]{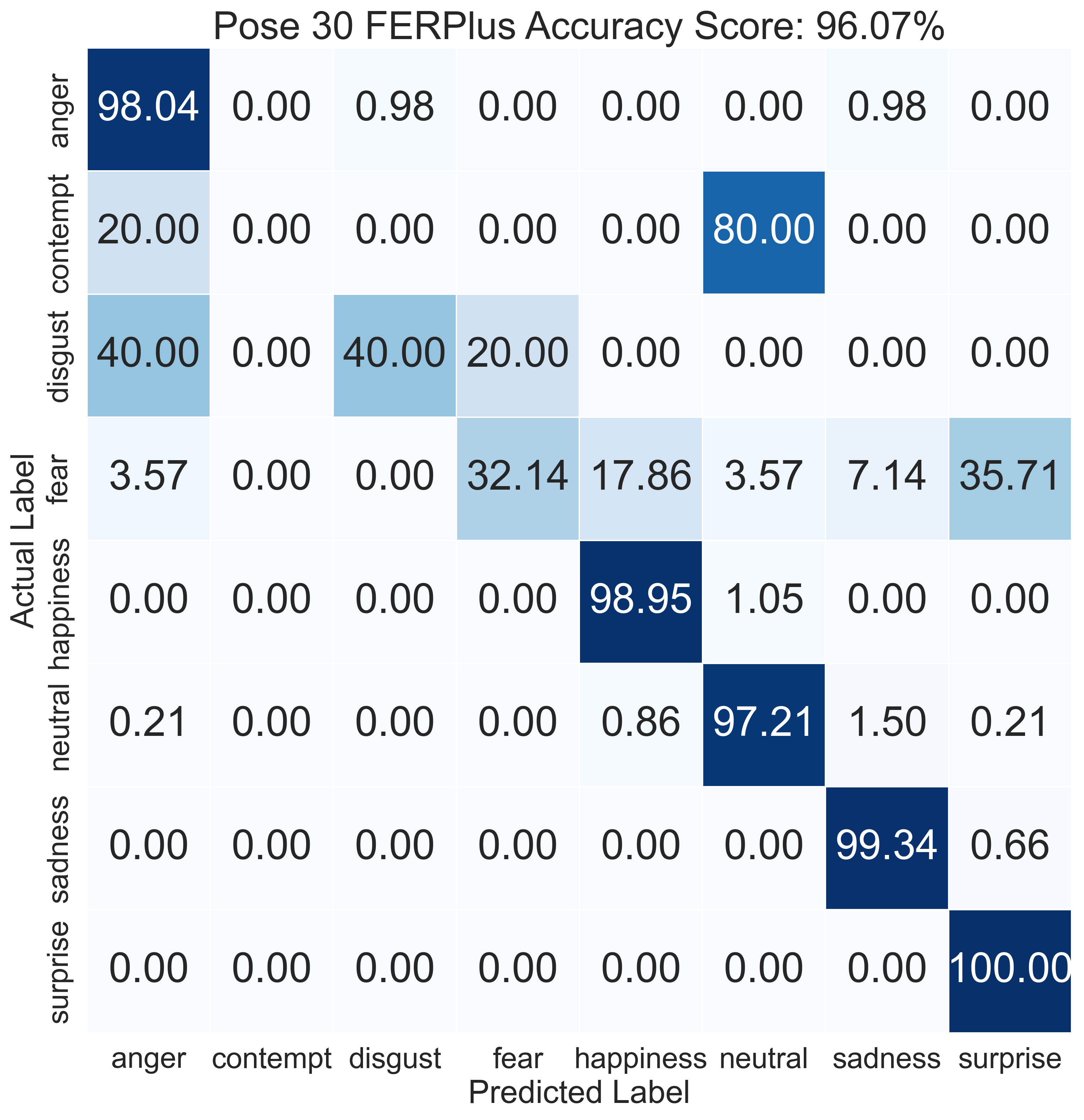}} \hfil
    \subfloat[Pose 45 FERPlus]{\includegraphics[width=2.25in]{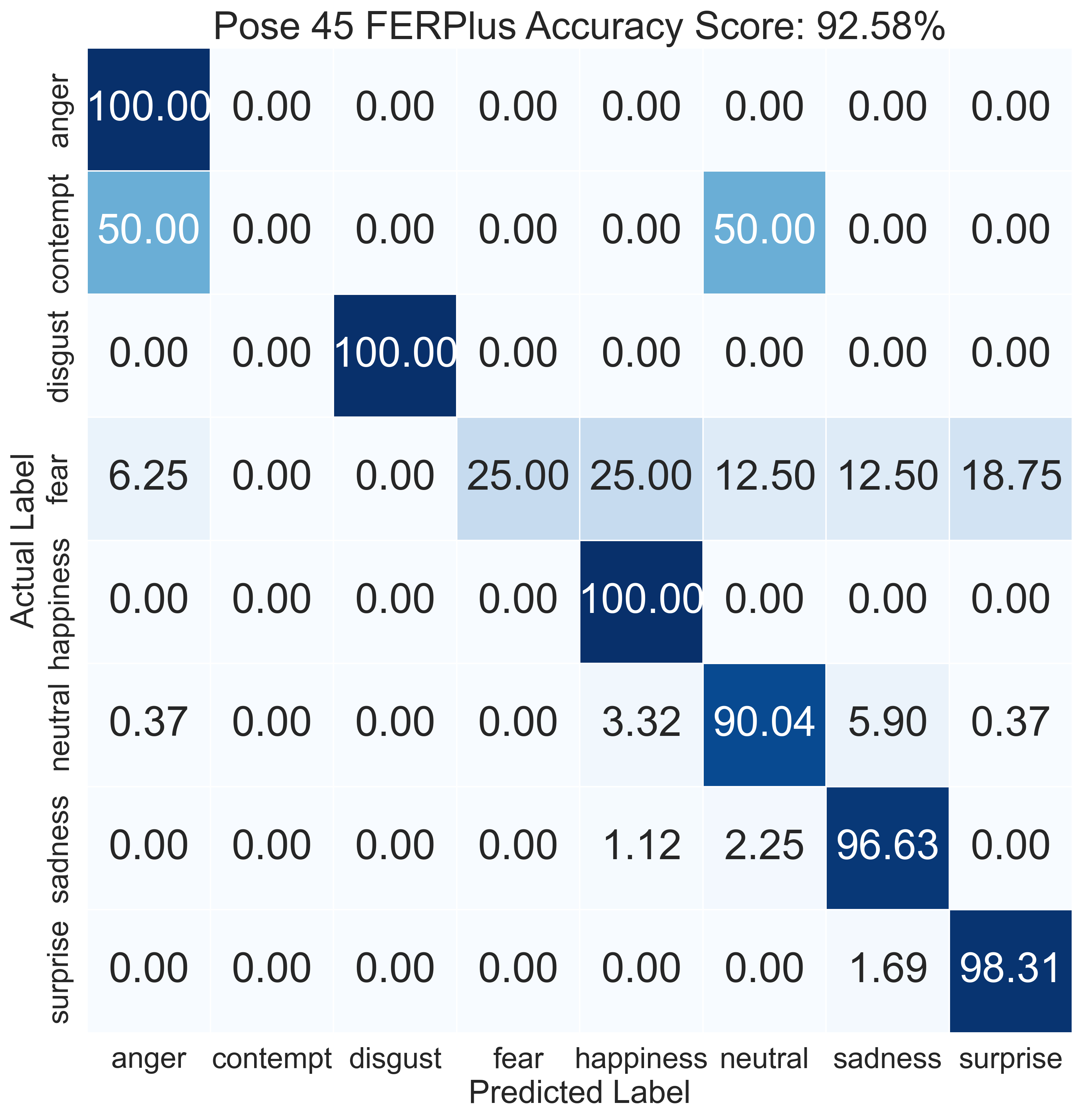}}
\caption{Confusion matrices of patch extraction on challenging subsets of FERPlus.}
\label{fig:extraction-ferp}
\end{figure*}

\begin{figure*}[!t]
\centering
    \subfloat[Occlusion FERPlus]{\includegraphics[width=2.25in]{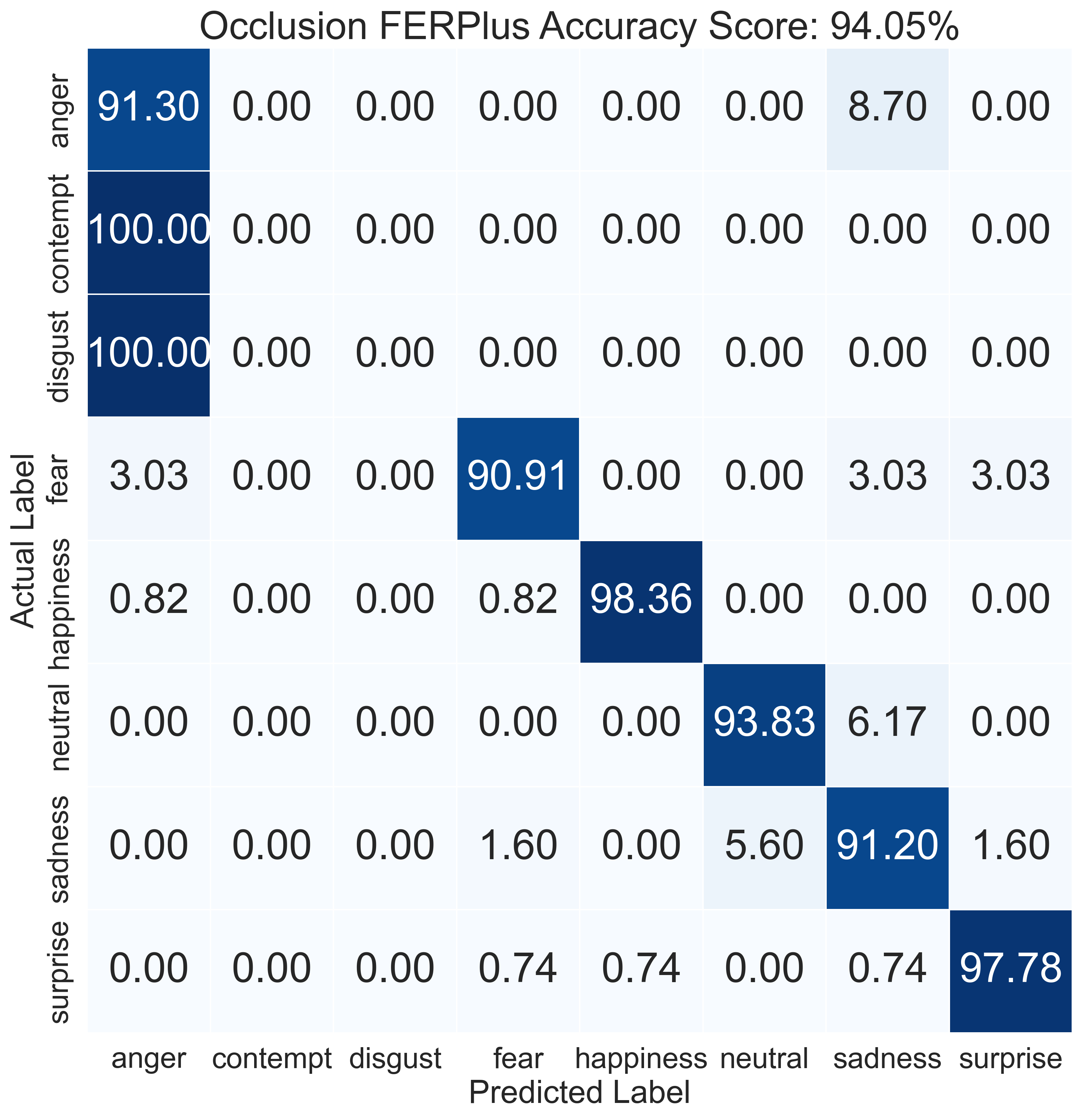}} \hfil
    \subfloat[Pose 30 FERPlus]{\includegraphics[width=2.25in]{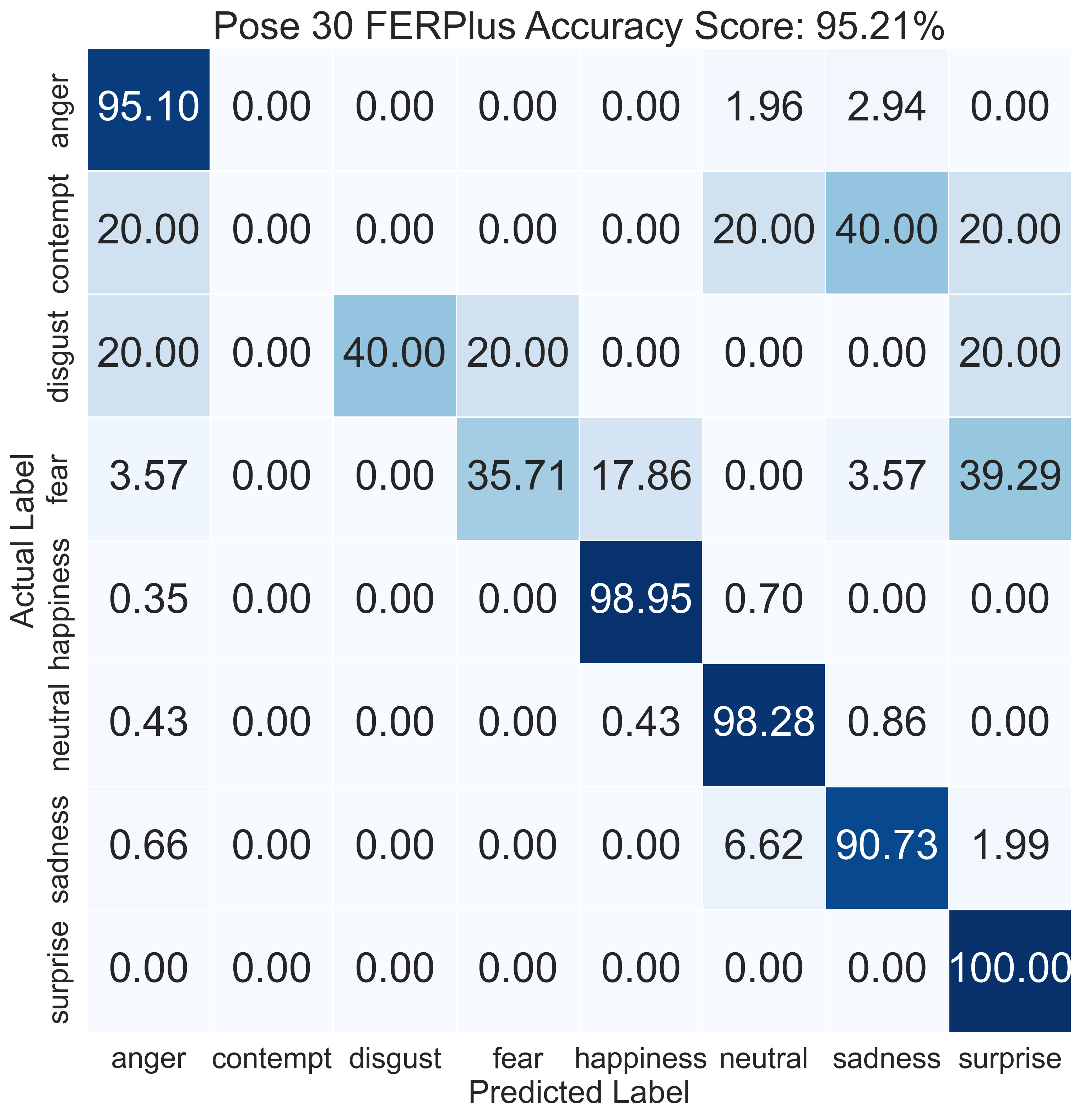}} \hfil
    \subfloat[Pose 45 FERPlus]{\includegraphics[width=2.25in]{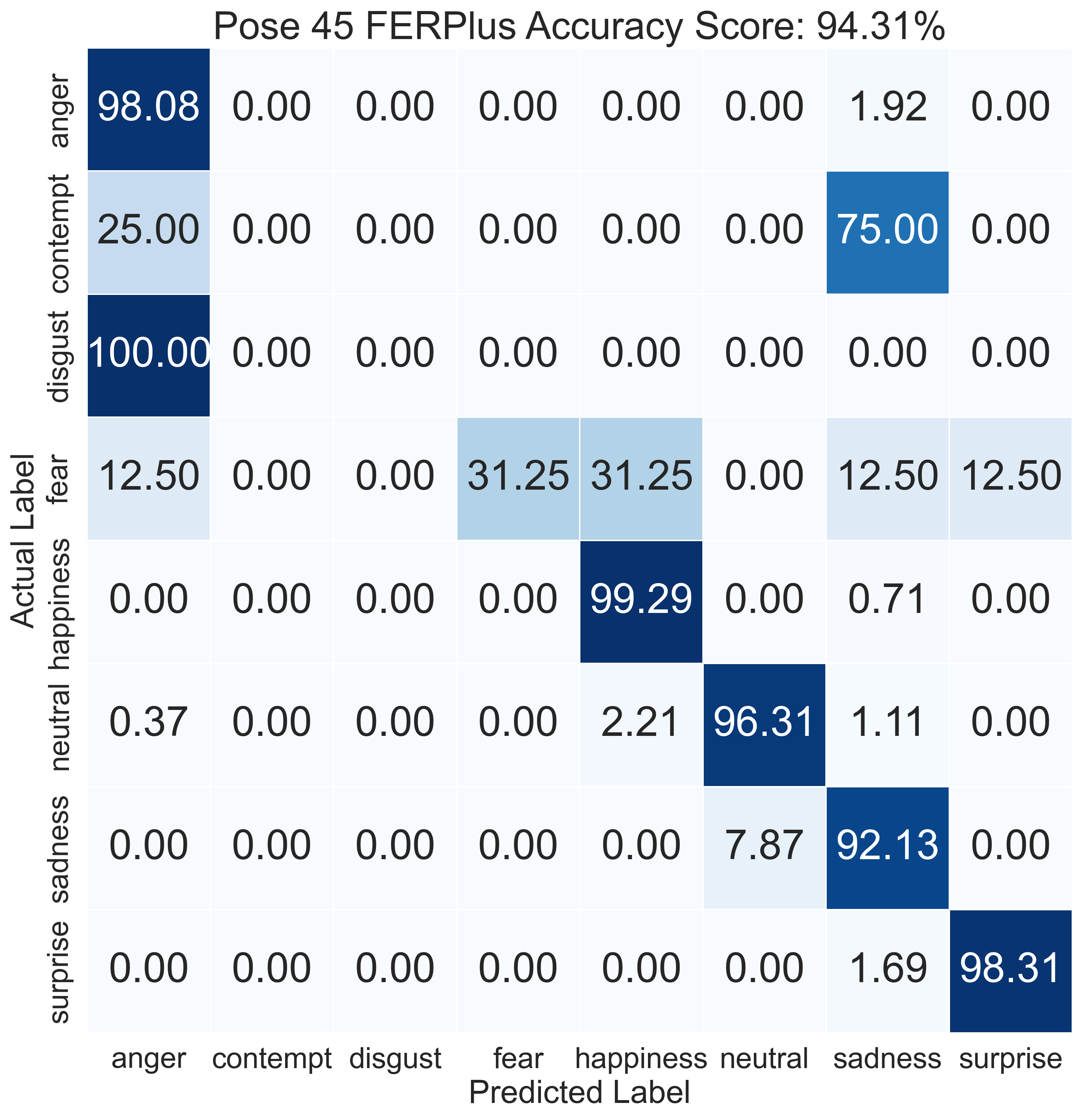}}
\caption{Confusion matrices of patch attention on challenging subsets of FERPlus.}
\label{fig:attention-ferp}
\end{figure*}

As shown in Table~\ref{tab:attention-extraction}, the conventional patch attention mechanism does not bring any performance gain when compared to the proposed patch extraction block. Instead, experimental results have demonstrated that the proposed patch extraction block performed better in the proposed method than the patch attention across all in-the-wild databases. 

Whereas the purpose of integrating a patch attention mechanism is often to highlight significant local facial regions and hence improve the classification performance under challenging conditions, the confusion matrices depicted in Fig. \ref{fig:extraction-rafdb} - \ref{fig:attention-ferp} have shown that the patch attention is identical to the patch extraction in terms of performance under challenging conditions. Specifically, the classification accuracy and per-class performance under challenging conditions are similar between these two designs. Both performed similarly well overall and in most classes except the Contempt class and the Disgust class, where both struggled to perform well. 

\subsection{Method Analysis}
\begin{figure*}
    \centering
    \includegraphics[width=\textwidth]{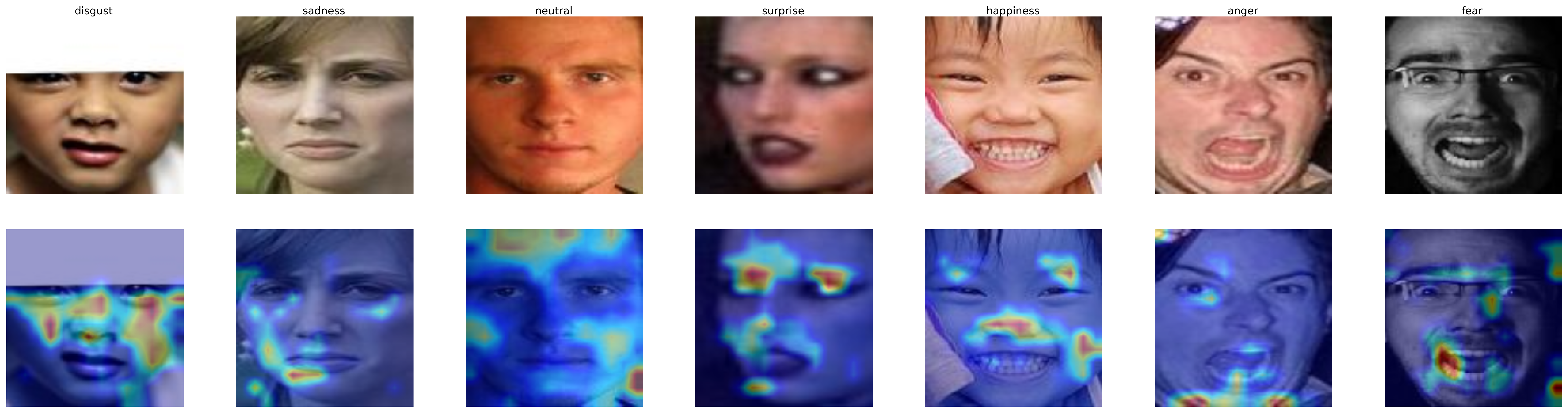}
    \caption{Grad-CAM of PAtt-Lite on all seven classes of RAF-DB. The first row is the sample images from the testing set whereas the second row is the Grad-CAM visualizations for each sample from the row above.}
    \label{fig:cam-classes}
\end{figure*}

\begin{figure*}
    \centering
    \includegraphics[width=\textwidth]{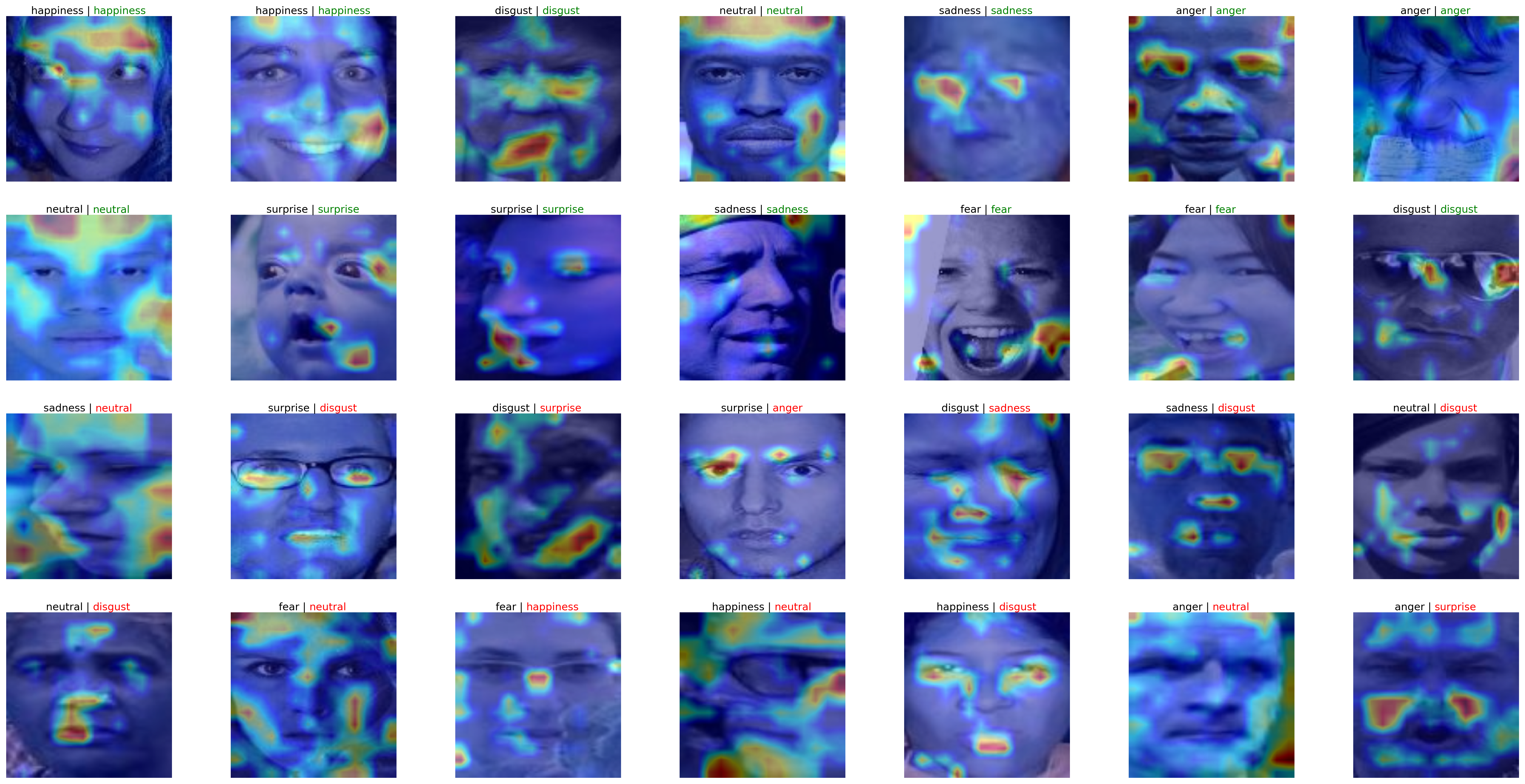}
    \caption{Examples of true and false predictions on RAF-DB and their activation maps. The ground truths of the samples are labeled in black whereas the true and false predictions are labeled in green and red, respectively.}
    \label{fig:cam-true-false}
\end{figure*}

In this section, PAtt-Lite is analyzed using Grad-CAM visualizations in Fig. \ref{fig:cam-classes} and Fig. \ref{fig:cam-true-false}. It can be observed that PAtt-Lite can recognize the facial expression of the subjects through or around the occlusion and under a variety of pose angles, besides some that have weird ratios. 

As highlighted in \cite{arnaud2022thin}, we believe expression annotations are the limiting factor of today's FER methods. Upon close inspection of the Grad-CAM from Fig. \ref{fig:cam-true-false}, it can be observed that some predictions, such as the fifth sample on the third row and the second sample on the fourth row, appear to be more closely represented by predictions from PAtt-Lite rather than the ground truths. These bad annotations are not limited to the test set but are available in the training set as well, which we believe contributed to PAtt-Lite's overfitting to the databases, as can be observed in Fig. \ref{fig:cam-true-false}, where some samples, such as the sixth samples on the second row, may seem ambiguous or incorrectly predicted. 

Besides the annotation, we also noticed that PAtt-Lite generally performed better on faces that are well-posed instead of faces that have a weird ratio, messy, or require context information. This is also illustrated in the visualizations in Fig. \ref{fig:cam-true-false}, such as the first sample on the third row and the fourth sample on the fourth row, where the backbone extractor randomly highlights the feature maps instead of focusing on the facial features. 

On the other hand, the lower class accuracy in smaller classes can be explained by referring to Table~\ref{tab:class-dist-wild}, Table~\ref{tab:class-dist-challenge}, and the confusion matrices, which most pointed that our method generally performed worse in the smaller classes. When the percentages are converted into the number of samples, it can be clearly observed that PAtt-Lite made a similar number of false predictions across all classes. 

\subsection{Method Comparison}
\begin{figure*}[!t]
\centering
    \subfloat[RAF-DB]{\includegraphics[width=2.25in]{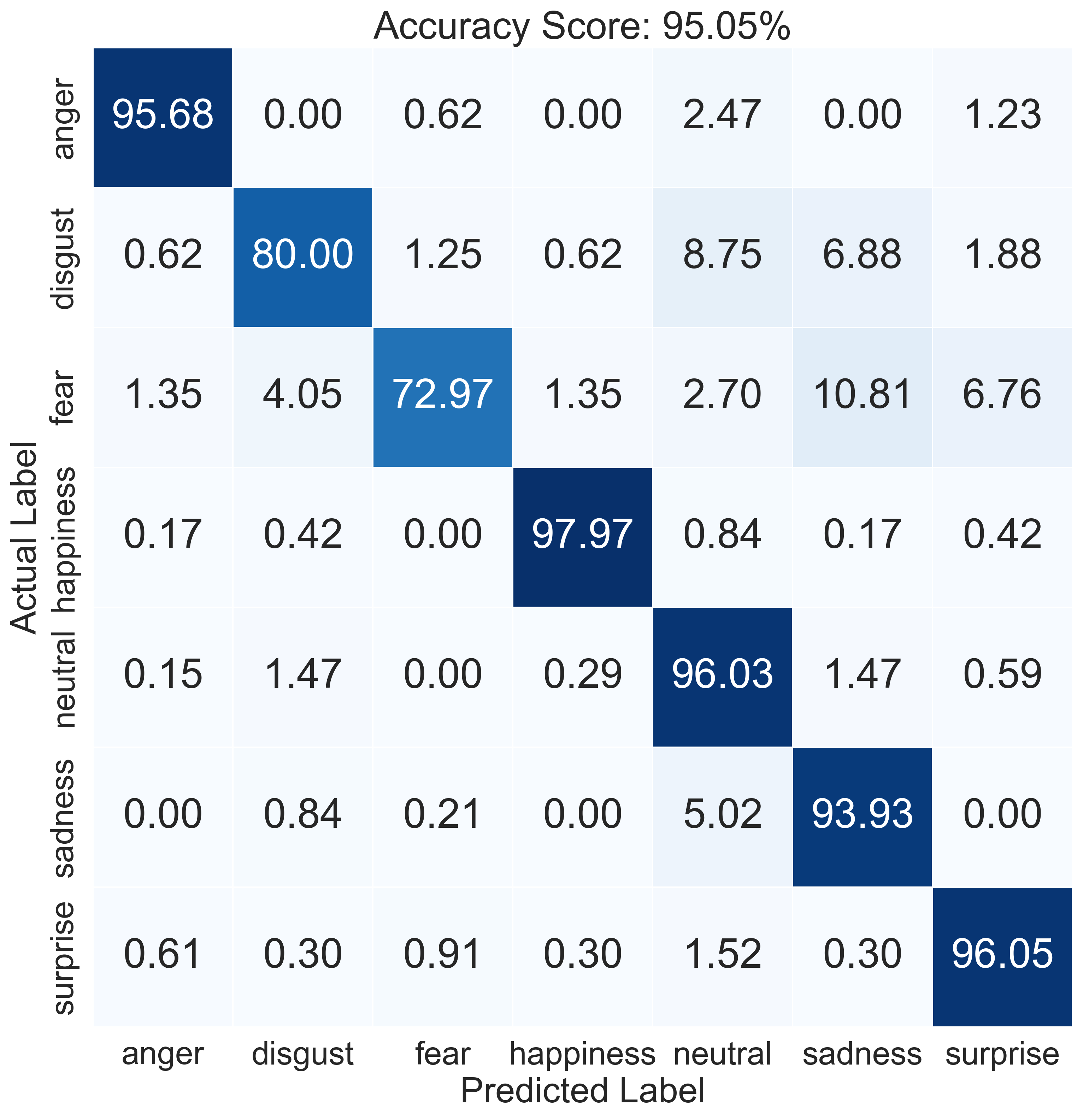}%
    \label{fig:rafdb-lite}}
    \hfil
    \subfloat[FER2013]{\includegraphics[width=2.25in]{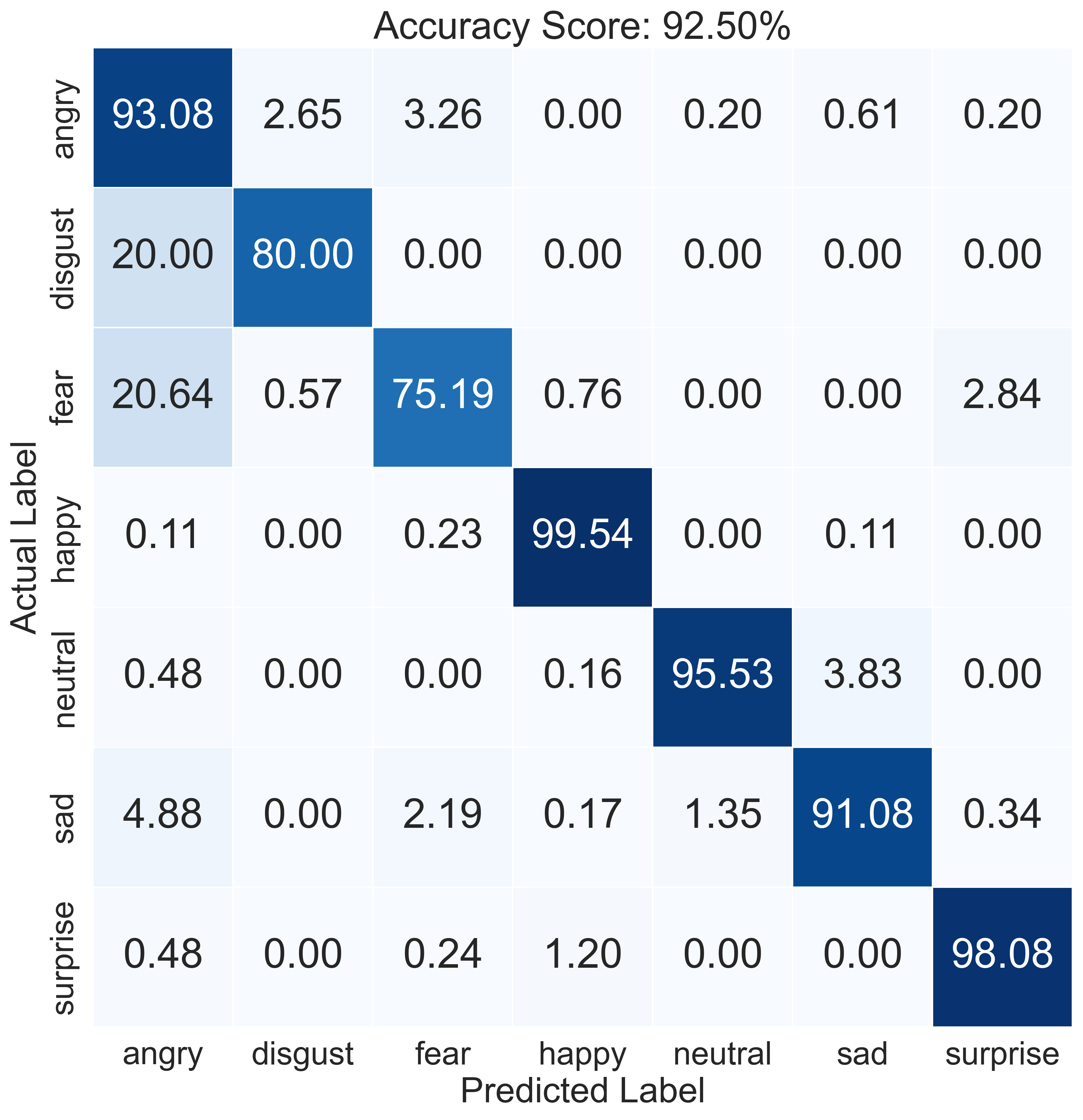}%
    \label{fig:fer13-lite}}
    \hfil
    \subfloat[FERPlus]{\includegraphics[width=2.25in]{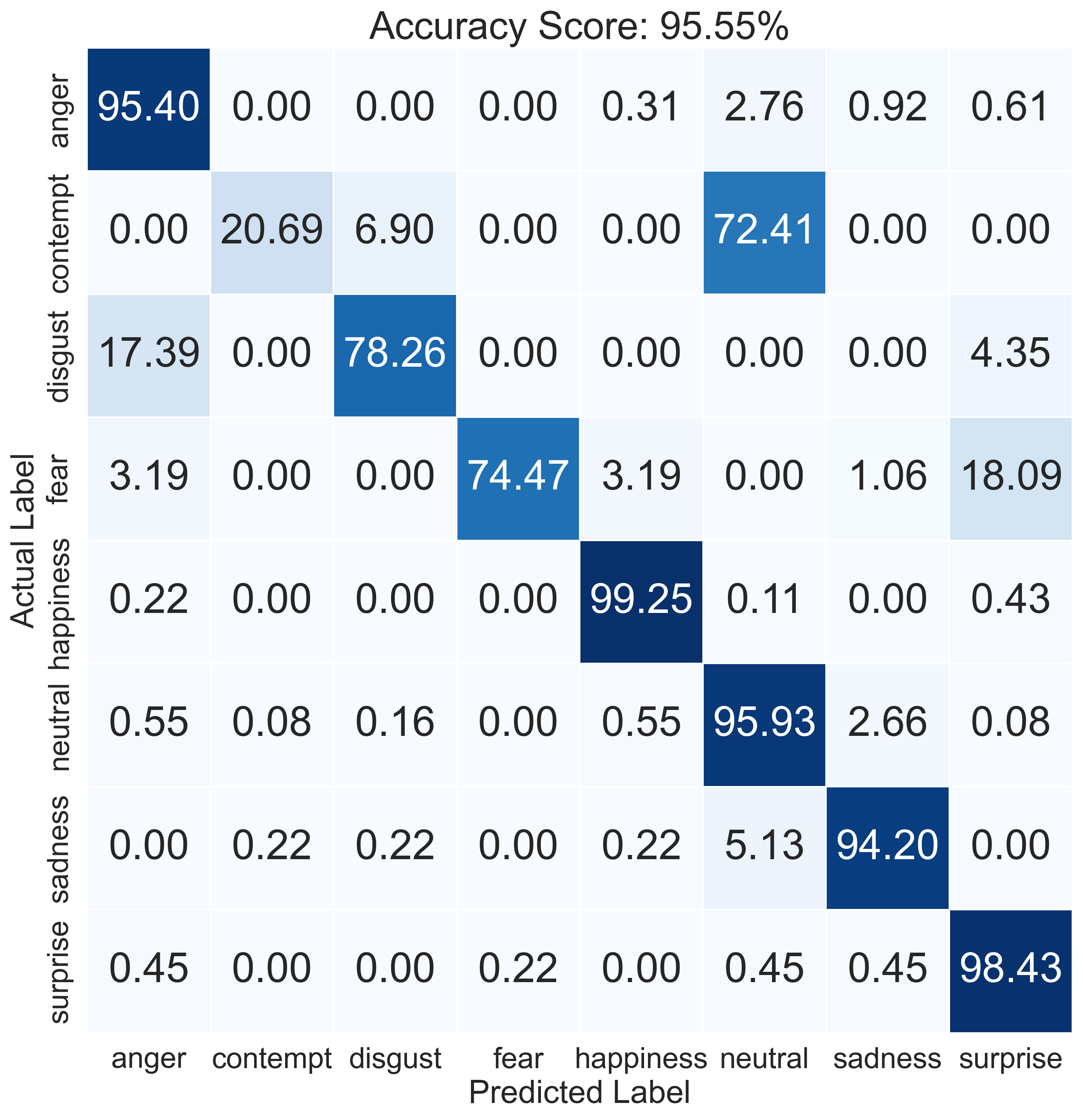}%
    \label{fig:ferp-lite}}
\caption{Confusion matrices of PAtt-Lite on in-the-wild databases.}
\label{fig:wild-lite}
\end{figure*}

In this section, PAtt-Lite is compared with state-of-the-art methods on the benchmark databases. The performance comparisons are shown in Table~\ref{tab:ckp-comparison} - \ref{tab:challenge-comparison}. In addition, the confusion matrices of the proposed PAtt-Lite on RAF-DB, FER2013, and FERPlus are depicted in Fig. \ref{fig:wild-lite}. 

\subsubsection{Results on CK+}
\begin{table}
    \caption{Comparison of the state-of-the-art results on CK+. The best result is highlighted in bold.}
    \label{tab:ckp-comparison}
    \centering
    \begin{tabular}{c c}
        \hline
        Methods & Accuracy \\
        \hline
        gACNN\cite{li2018occlusion} & 96.40 \\
        pACNN\cite{li2018occlusion} & 97.03 \\
        SCAN-CCI\cite{gera2021landmark} & 97.31 \\
        IF-GAN\cite{cai2021identity} & 97.52 \\
        FDRL\cite{ruan2021feature} & 99.54 \\
        ViT + SE\cite{aouayeb2021learning} & 99.80 \\
        \textbf{FER-VT\cite{huang2021facial}} &\textbf{100.00}\\
        \hline
        \textbf{PAtt-Lite} & \textbf{100.00} \\
        \hline
    \end{tabular}
\end{table}

The performance comparison of the proposed method with the state-of-the-art methods for CK+ is presented in Table~\ref{tab:ckp-comparison}. The proposed PAtt-Lite outperformed all CNN-based existing work\cite{li2018occlusion, gera2021landmark, cai2021identity, ruan2021feature} and transformer-based existing work\cite{aouayeb2021learning} in terms of cross-validation mean accuracy for CK+ by achieving 100.00\% mean accuracy across 10-fold cross-validation. To the best of our knowledge, FER-VT\cite{huang2021facial} is the only other method that reported the same performance with a transformer-based method. 

\subsubsection{Results on RAF-DB.}
\begin{table}
    \caption{Comparison of the state-of-the-art results on RAF-DB and FERPlus. The best result is highlighted in bold.}
    \label{tab:rafdb-ferp-comparison}
    \centering
    \begin{tabular}{c c c c}
        \hline
        Methods                         & \# Params & RAF-DB & FERPlus \\
        \hline
        VTFF\cite{ma2021facial}         & 80.1M     & - & 88.81 \\
        RAN\cite{wang2020region}        & 11.2M     & 86.90 & 89.16 \\
        VTFF\cite{ma2021facial}         & 51.8M     & 88.14 & - \\
        SCAN-CCI\cite{gera2021landmark} & 70M       & 89.02 & 89.42 \\
        EAC\cite{zhang2022learn}        & 11.2M     & 89.99 & 89.64 \\
        ARM\cite{shi2021learning}       & 11.2M     & 90.42 & - \\
        TransFER\cite{xue2021transfer}  & 65.2M     & 90.91 & 90.83 \\
        Facial Chirality\cite{lo2022facial} & 46.2M     & 91.20 & - \\
        DDAMFN\cite{zhang2023dual}      & 4.11M     & 91.35 & 90.74 \\
        APViT\cite{xue2022vision}       & 65.2M     & 91.98 & 90.86 \\
        POSTER\cite{zheng2022poster}    & 71.8M     & 92.05 & 91.62 \\
        POSTER++\cite{mao2023poster}    & 43.7M     & 92.21 & - \\
        ARBEx\cite{wasi2023arbex}       & -         & 92.47 & 93.09 \\
        CIAO\cite{barros2022ciao}       & 17.9M     & - & 94.50 \\
        \hline
        \textbf{PAtt-Lite} & \textbf{1.10M} & \textbf{95.05} & \textbf{95.55} \\
        \hline
    \end{tabular}
\end{table}

\begin{table*}
    \caption{Per-class performance comparison of the state-of-the-art results on RAF-DB. The best result is highlighted in bold.}
    \label{tab:rafdb-class-comparison}
    \centering
    \begin{tabular}{c c c c c c c c c c}
        \hline
        Method                                    & \# Params & Anger           & Disgust         & Fear            & Happiness       & Neutral         & Sadness         & Surprise        & Average         \\
        \hline
        Imponderous Net\cite{gera2021imponderous} & 1.45M     & 78.00           & 54.00           & 57.00           & 96.00           & 88.00           & 85.00           & 85.00           & 77.57           \\
        MViT\cite{li2021mvt}                      & 33M       & 78.40           & 63.75           & 60.81           & 95.61           & 89.12           & 87.45           & 87.54           & 80.38           \\
        VTFF\cite{ma2021facial}                   & 51.8M     & 85.80           & 68.12           & 64.86           & 94.09           & 87.50           & 87.24           & 85.41           & 81.86           \\
        SCAN-CCI\cite{gera2021landmark}           & 70M       & 81.00           & 70.00           & 66.00           & 96.00           & 89.00           & 86.00           & 88.00           & 82.29           \\
        ARM\cite{shi2021learning}                 & 11.2M     & 77.20           & 64.40           & 70.30           & 95.40           & \textbf{97.90}  & 83.90           & 90.30           & 82.77           \\
        TransFER\cite{xue2021transfer}            & 65.2M     & 88.89           & 79.37           & 68.92           & 95.95           & 90.15           & 88.70           & 89.06           & 85.86           \\
        POSTER++\cite{mao2023poster}              & 43.7M     & 88.27           & 71.88           & 68.92           & 97.22           & 92.06           & 92.89           & 90.58           & 85.97           \\
        POSTER\cite{zheng2022poster}              & 71.8M     & 88.89           & 75.00           & 67.57           & 96.96           & 92.35           & 91.21           & 90.27           & 86.04           \\
        APViT\cite{xue2022vision}                 & 65.2M     & 86.42           & 73.75           & \textbf{72.97}  & 97.30           & 92.06           & 88.70           & 93.31           & 86.36           \\
        \hline
        \textbf{PAtt-Lite}                        & \textbf{1.10M} & \textbf{95.68}  & \textbf{80.00}  & \textbf{72.97}  & \textbf{97.97}  & 96.03 & \textbf{93.93}  & \textbf{96.05}  & \textbf{90.38}  \\
        \hline
    \end{tabular}
\end{table*}

\begin{figure}
    \centering
    \includegraphics[width=\columnwidth]{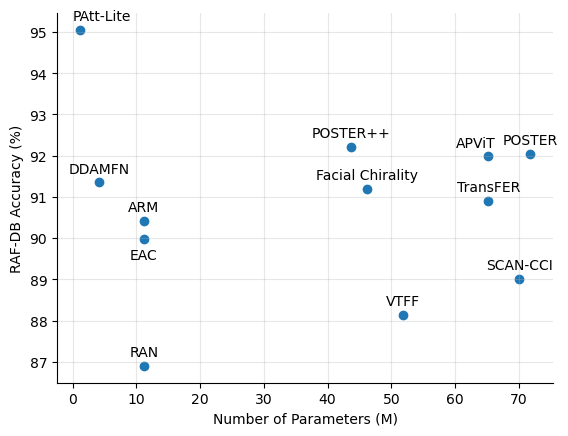}
    \caption{Accuracy and parameter comparison on RAF-DB.}
    \label{fig:rafdb-params}
\end{figure}

The performance comparison of the proposed PAtt-Lite with state-of-the-art methods on RAF-DB is shown in Table~\ref{tab:rafdb-ferp-comparison}. An accuracy and parameter comparison between the proposed method and state-of-the-art methods is also presented in Fig. \ref{fig:rafdb-params}. VTFF\cite{ma2021facial}, TransFER\cite{xue2021transfer}, Facial Chirality\cite{lo2022facial}, APViT\cite{xue2022vision}, POSTER\cite{zheng2022poster}, POSTER++\cite{mao2023poster}, and ARBEx\cite{wasi2023arbex} are transformer-based architecture, whereas RAN\cite{wang2020region}, SCAN-CCI\cite{gera2021landmark}, ARM\cite{shi2021learning}, EAC\cite{zhang2022learn}, and DDAMFN\cite{zhang2023dual} are CNN-based architecture. 

Based on the comparison, the transformer-based methods are generally outperforming and have a greater number of parameters than the CNN-based methods, with SCAN-CCI\cite{gera2021landmark} being the exception as it has 70M parameters despite having a CNN-based architecture. Our proposed PAtt-Lite achieved better performance with a CNN backbone than all transformer-based state-of-the-art. Specifically, PAtt-Lite has 2.58\% over transformer-based ARBEx\cite{wasi2023arbex}, a modified version of the lightest transformer-based method, POSTER++\cite{mao2023poster}, recorded the best result for RAF-DB. For comparison with CNN-based methods, PAtt-Lite achieved an improvement of 4.63\% overall accuracy over ARM\cite{shi2021learning}. In terms of performance comparison with lightweight methods, our proposed method also outperformed RAN\cite{wang2020region}, EAC\cite{zhang2022learn}, ARM\cite{shi2021learning}, and DDAMFN\cite{zhang2023dual} by 8.15\%, 5.06\%, 4.63\%, and 3.70\% respectively, while having significantly lesser parameters. 

Furthermore, following \cite{ma2021facial, zheng2022poster, xue2022vision, mao2023poster}, a per-class performance comparison is added in Table~\ref{tab:rafdb-class-comparison} to compare the performance of PAtt-Lite on different classes of the database. For \cite{gera2021landmark, gera2021imponderous, li2021mvt, xue2021transfer, shi2021learning}, which did not specifically report their per-class accuracy, their results from the confusion matrices are taken for comparison. Overall, a similar trend for the per-class performance is observed when PAtt-Lite is compared with existing work, where the per-class accuracy for the Disgust class and the Fear class is lacking behind the other classes. However, our method managed to slightly improve the per-class accuracy for the Disgust class over APViT\cite{xue2022vision} and perform on par with it for the Fear class while having 59 times fewer parameters. Overall, small improvements can be observed across all expression classes except class Anger and class Neutral, where significant improvement is recorded for the former and a small performance drop is noted for the latter. Noticeably, the proposed method managed to record a 6.79\% improvement over all other state-of-the-art methods for class Anger, while recording a 1.87\% performance drop over ARM\cite{shi2021learning}. Although the per-class accuracy for the Disgust class and the Fear class can be further improved, PAtt-Lite demonstrated consistently strong performance by managing to achieve about 93\% per-class accuracy or greater in the remaining classes, resulting in an average accuracy of 90.38\%. This corresponds to an improvement in the average accuracy of 4.02\% over APViT\cite{xue2022vision}, which previously reported the best average accuracy. Moreover, when compared to Imponderous Net\cite{gera2021imponderous}, which has the nearest number of parameters as our proposed method, PAtt-Lite achieved an improvement of 12.81\% in terms of average accuracy, while performing stronger and more consistent on all expression classes. 

\subsubsection{Results on FER2013}
\begin{table}
    \caption{Comparison of the state-of-the-art results on FER2013. The best result is highlighted in bold.}
    \label{tab:fer13-comparison}
    \centering
    \begin{tabular}{c c}
        \hline
        Methods & Accuracy \\
        \hline
        KDL\cite{mahmoudi2020kernelized} & 71.28 \\
        MAFER\cite{massoli2021mafer} & 73.45 \\
        PASM\cite{liu2021point} & 73.59 \\
        LHC-Net\cite{pecoraro2022local} & 74.42 \\
        EmoNeXt-XLarge\cite{el2023emonext} & 76.12 \\
        MoVE-CNNs\cite{yu2021move} & 77.70 \\
        FLEPNet\cite{karnati2022flepnet} & 80.72 \\
        NECM-PECM Ensemble\cite{phattarasooksirot2022facial} & 88.00 \\
        \hline
        \textbf{PAtt-Lite} & \textbf{92.50} \\
        \hline
    \end{tabular}
\end{table}

The comparison of the proposed method with the state-of-the-art methods on FER2013 is depicted in Table~\ref{tab:fer13-comparison}. FER2013 has been a database that existing work struggled to perform well on until recently when FLEPNet\cite{karnati2022flepnet} and NECM-PECM Ensemble\cite{phattarasooksirot2022facial} were proposed. Compared to the existing work, PAtt-Lite recorded a classification performance of 92.50\%, which equates to an improvement of 4.5\% over NECM-PECM Ensemble\cite{phattarasooksirot2022facial}. 

From the confusion matrix in Fig \ref{fig:fer13-lite}, the per-class performance of our proposed method on FER2013 is observed to have a similar trend to RAF-DB and existing work\cite{karnati2022flepnet, massoli2021mafer}, where generally strong per-class accuracy can be achieved for all expression classes except the Disgust class and the Fear class, which are relatively weaker compared to the remaining classes. However, PAtt-Lite has managed to achieve significantly better performance across all expression classes. Specifically, the proposed method improves the per-class accuracy of all expressions except Disgust and Fear to more than 90\% accuracy. This is unlike most existing work, which has struggled to achieve such performance on all expressions except the Happy class. 

\subsubsection{Results on FERPlus}
\begin{figure}
    \centering
    \includegraphics[width=\columnwidth]{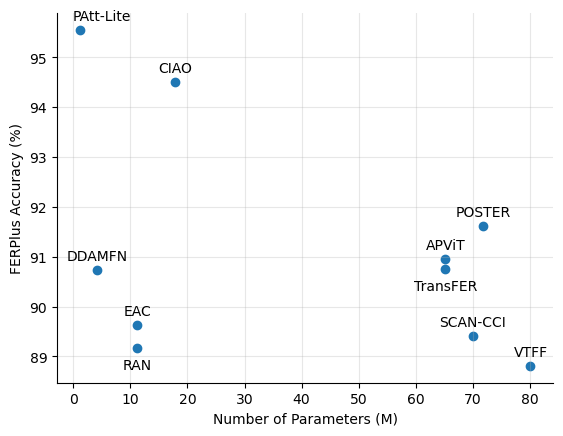}
    \caption{Accuracy and parameter comparison on FERPlus.}
    \label{fig:ferp-params}
\end{figure}

FERPlus has an additional Contempt expression than the other in-the-wild databases in this research. The classification performance of our proposed method is compared with the state-of-the-art methods in the last column of Table~\ref{tab:rafdb-ferp-comparison}. The accuracy and parameter of the proposed PAtt-Lite are also compared with state-of-the-art methods in Fig. \ref{fig:ferp-params}. Like RAF-DB, transformer-based methods generally have better results than CNN-based methods, except for CIAO\cite{barros2022ciao}, which reported the best accuracy for FERPlus with a CNN backbone. 

The proposed PAtt-Lite has achieved slight improvement on FERPlus when compared to the state-of-the-art methods. Specifically, PAtt-Lite achieved a classification accuracy of 95.55\%, which corresponds to a performance improvement of 1.05\% over CIAO\cite{barros2022ciao}, while having 16 times lesser parameters than CIAO\cite{barros2022ciao}. When compared to transformer-based methods, our method achieved a 3.93\% improvement in classification accuracy over POSTER\cite{zheng2022poster} with 65 times lesser parameters, and a 2.46\% improvement over ARBEx\cite{wasi2023arbex}, a modified POSTER++\cite{mao2023poster}. For comparison with lighter state-of-the-art methods, PAtt-Lite with significantly lesser parameters also outperformed RAN\cite{wang2020region}, EAC\cite{zhang2022learn}, and DDAMFN\cite{zhang2023dual} by 6.39\%, 5.91\%, and 4.81\%, respectively, in terms of overall accuracy. 

It is visible that our proposed method is still struggling with the Contempt class upon inspecting the confusion matrix in Fig. \ref{fig:ferp-lite}, with the Disgust class and the Fear class following a general trend from RAF-DB and FER2013. However, this performance drop is expected due to the severe class imbalance between these classes and the larger class. Similar performance drops on some of these negative expression classes are also reported in existing work\cite{gera2021landmark, xue2022vision, zheng2022poster}. Despite this, our proposed method still achieved significant improvements in all other expression classes, with around or higher than 95\% performance across these 5 classes. 

\subsubsection{Results on challenging subsets}
\begin{table*}[!ht]
    \caption{Comparison of the state-of-the-art results on challenging subsets for RAF-DB and FERPlus. The best result is highlighted in bold.}
    \label{tab:challenge-comparison}
    \centering
    \begin{tabular}{c c c c c c c c}
    \hline
    \multirow{2}{*}{Methods} & \multirow{2}{*}{\# Params} & \multicolumn{3}{c}{RAF-DB} & \multicolumn{3}{c}{FERPlus} \\
    \cline{3-8}
            & & Occlusion & Pose $>$ 30 & Pose $>$ 45 & Occlusion & Pose $>$ 30 & Pose $>$ 45 \\
    \hline
    RAN\cite{wang2020region}                                & 11.2M     & 82.72 & 86.74 & 85.20 & 83.63 & 82.23 & 80.40 \\
    Imponderous Net\cite{gera2021imponderous}               & 1.45M     & 83.40 & 86.12 & 84.41 & 83.47 & 86.84 & 84.83 \\
    VTFF\cite{ma2021facial}                                 & 51.8M     & 83.95 & 87.97 & 88.35 & - & - & - \\
    VTFF\cite{ma2021facial}                                 & 80.1M     & - & - & - & 84.79 & 88.29 & 87.20 \\
    OADN\cite{ding2020occlusion}                            & -         & - & - & - & 84.57 & 88.52 & 87.50 \\
    SCAN-CCI\cite{gera2021landmark}                         & 70M       & 85.03 & 89.82 & 89.07 & 86.12 & 88.89 & 88.15 \\
    MViT\cite{li2021mvt}                                    & 33M       & 85.17 & 87.99 & 88.40 & - & - & - \\
    Facial Chirality\cite{lo2022facial}                     & 46.2M     & 88.16 & 91.50 & 90.86 & - & - & - \\
    \hline
    \textbf{PAtt-Lite} & \textbf{1.10M} & \textbf{92.23} & \textbf{95.35} & \textbf{94.44} & \textbf{93.22} & \textbf{96.07} & \textbf{92.58} \\
    \hline
    \end{tabular}
\end{table*}

The proposed method has shown great results in classifying samples under challenging conditions. Table~\ref{tab:challenge-comparison} shows the difference in the performance of the proposed method compared to recent work on challenging subsets for RAF-DB and FERPlus. Specifically, PAtt-Lite achieved an improvement of more than 3\% across all three challenging subsets for RAF-DB, outperforming Facial Chirality\cite{lo2022facial} which reported the best results on these subsets with 46.2M parameters. Meanwhile, to our best knowledge, our proposed method is the first to achieve more than 90\% accuracy for all challenging subsets for FERPlus. A performance improvement of around 4\% to 7\% is achieved across the three subsets compared to SCAN-CCI\cite{gera2021landmark}, which previously reported the best performance with 70M parameters. From the comparison with the best results on these subsets, the proposed PAtt-Lite achieved state-of-the-art performance across all subsets with significantly lesser parameters. Meanwhile, when compared with the lighter methods such as RAN\cite{wang2020region} and Imponderous Net\cite{gera2021imponderous} on subsets of RAF-DB, PAtt-Lite managed to outperform these methods by around 9\%. The proposed method also managed to outperform the lighter methods by more than 9\% on the first two subsets and by 7.75\% on the Pose 45 subset of FERPlus. 

\section{Conclusion}
This work presents PAtt-Lite, a MobileNetV1-based solution, to improve the classification accuracy of FER under challenging conditions. The proposed PAtt-Lite achieves state-of-the-art performance in all benchmark databases and their subsets while being significantly lighter than other state-of-the-art methods at just 1.10M parameters. Specifically, the proposed patch extraction block improves the FER performance of PAtt-Lite under challenging conditions by enforcing the model to extract significant local facial features. On the other hand, the attention classifier is proposed to learn the patched representation better and improve the overall performance of the proposed lightweight method. 

This work provides valuable insight into potential future directions by highlighting the advantages and possible improvements of our proposed method. One such direction is robustness improvement to PAtt-Lite, especially towards low-resource expressions, which can be done by further refining the patch extraction blocks or experimenting with a feature extractor backbone that has more performance or is more robust. Besides, as highlighted in the previous section, while large FER databases are available, good annotations are required to further advance the field of FER. Hence, we also suggest that there could be an effort for the development of a FER database with more reliable annotations. Existing FER methods have advanced to a point where they can potentially discover patterns that are difficult for human annotators to detect. Therefore, a crowdsourced database that incorporates insights from state-of-the-art FER methods could be a valuable resource for further research in this field. 

Overall, this work has shown the potential to use MobileNetV1 as a baseline feature extractor in FER. The need for continued research and development is also highlighted to further improve the accuracy and reliability of automated facial expression recognition, especially under challenging conditions and low-resource expressions. 

\bibliographystyle{unsrt}
\bibliography{refs}

\begin{IEEEbiography}[{\includegraphics[width=1in,height=1.25in,clip,keepaspectratio]{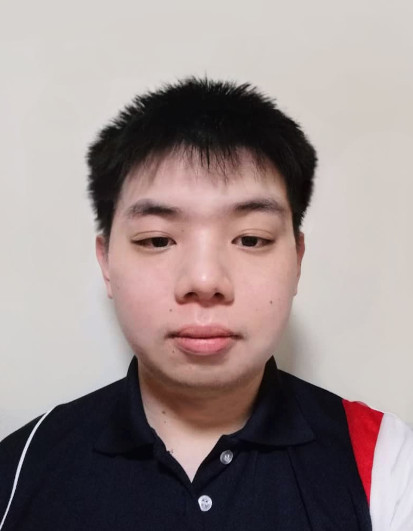}}]{Jia Le Ngwe} received the bachelor's degree in Artificial Intelligence (AI) from the Multimedia University, Malaysia, in 2022. He is currently pursuing his Ph.D. degree in the field of AI at Multimedia University. His research interest includes generative modelling, affective computing, and edge AI.
\end{IEEEbiography}

\begin{IEEEbiography}[{\includegraphics[width=1in,height=1.25in,clip,keepaspectratio]{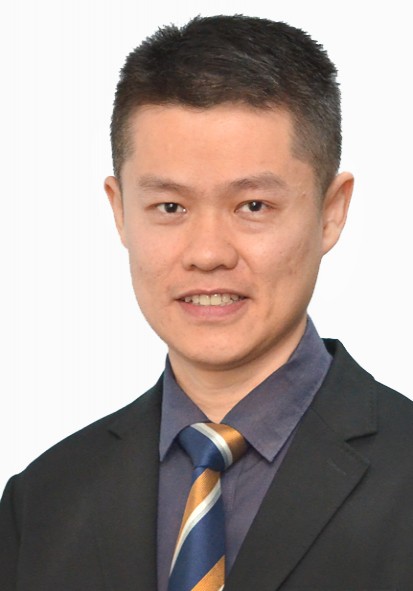}}]{Kian Ming Lim} received B.IT (Hons) in Information Systems Engineering, Master of Engineering Science (MEngSc) and Ph.D. (I.T.) degrees from Multimedia University. He is currently a Lecturer with the Faculty of Information Science and Technology, Multimedia University. His research and teaching interests includes machine learning, deep learning, and computer vision and pattern recognition.
\end{IEEEbiography}

\begin{IEEEbiography}[{\includegraphics[width=1in,height=1.25in,clip,keepaspectratio]{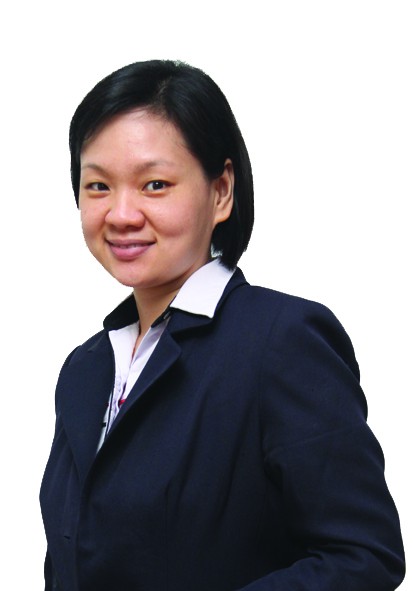}}]{Chin Poo Lee} is a Senior Lecturer in the Faculty of Information Science and Technology at Multimedia University, Malaysia. She completed her Masters of Science and Ph.D. in Information Technology in the area of abnormal behaviour detection and gait recognition. She is a certified Professional Technologist since 2018, a member of International Association of Engineers since 2020 as well as Outcome-Based Education Consultant and Trainer. Her research interests include action recognition, computer vision, gait recognition, natural language processing and deep learning.
\end{IEEEbiography}

\begin{IEEEbiography}[{\includegraphics[width=1in,height=1.25in,clip,keepaspectratio]{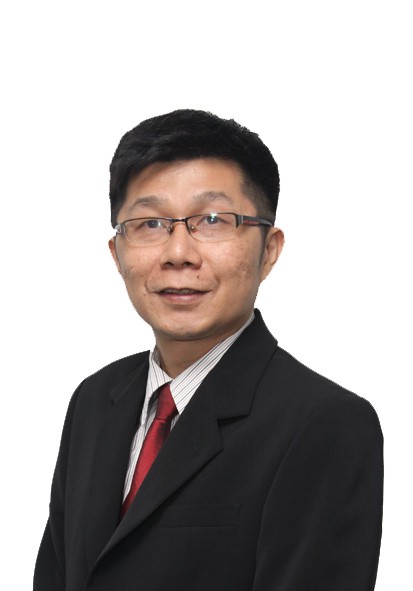}}]{Thian Song Ong} received the M.Sc. degree from the University of Sunderland, U.K., in 2001, and the Ph.D. degree from Multimedia University, Malaysia, in 2008. He works with the Faculty of Information Sciences and Technology (FIST), Multimedia University. He has more than 70 publications from conferences and international refereed journals published. His research interests include biometric security and machine learning. From 2013 to 2015, he served on the Editorial Board of the IEEE BIOMETRICS COUNCIL NEWSLETTER.
\end{IEEEbiography}

\begin{IEEEbiography}[{\includegraphics[width=1in,height=1.25in,clip,keepaspectratio]{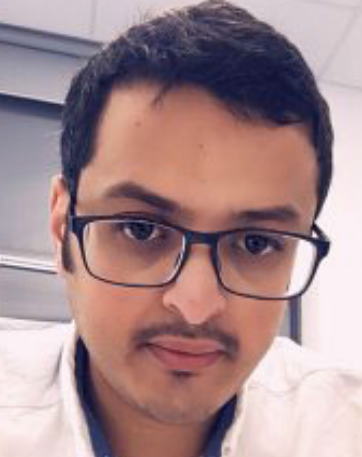}}]{Ali Alqahtani} received the Ph.D. degree in computer science from Swansea University, Swansea, U.K., in 2021. He is currently an Assistant Professor with the Department of Computer Science, King Khalid University, Abha, Saudi Arabia. He has published several refereed conference and journal publications. His research interests include various aspects of pattern recognition, deep learning, and machine intelligence and their applications to real-world problems.
\end{IEEEbiography}

\EOD

\end{document}